\PassOptionsToPackage{svgnames}{xcolor}
\documentclass[sigconf]{acmart}

\AtBeginDocument{%
  \providecommand\BibTeX{{%
    \normalfont B\kern-0.5em{\scshape i\kern-0.25em b}\kern-0.8em\TeX}}}

\setcopyright{none}
\acmConference[FAccT '22]{2022 ACM Conference on Fairness, Accountability, and Transparency}{June 21--24, 2022}{Seoul, Republic of Korea}
\acmBooktitle{2022 ACM Conference on Fairness, Accountability, and Transparency (FAccT '22), June 21--24, 2022, Seoul, Republic of Korea}
\acmDOI{10.1145/3531146.3533184}
\acmISBN{978-1-4503-9352-2/22/06}
\usepackage{subfig}
\usepackage{comment}
\usepackage[svgnames]{xcolor}
\usepackage{amsmath}
\newcommand{\etal}{\textit{et al}. }
\newcommand{\ie}{\textit{i}.\textit{e}., }
\newcommand{\eg}{\textit{e}.\textit{g}., }

\makeatletter
\newcommand{\figcaption}[1]{\def\@captype{figure}\caption{#1}}
\newcommand{\tblcaption}[1]{\def\@captype{table}\caption{#1}}
\makeatother

\acmISBN{978-1-4503-9352-2/22/06}

\begin{document}

\title{Gender and Racial Bias in Visual Question Answering Datasets}

\author{Yusuke Hirota}
\email{y-hirota@is.ids.osaka-u.ac.jp}
\orcid{0000-0002-7221-0184}
\affiliation{%
  \institution{Osaka University}
  \city{Osaka}
  \country{Japan}
}

\author{Yuta Nakashima}
\email{n-yuta@ids.osaka-u.ac.jp}
\orcid{0000-0001-8000-3567}
\affiliation{%
  \institution{Osaka University}
  \city{Osaka}
  \country{Japan}
}

\author{Noa Garcia}
\email{noagarcia@ids.osaka-u.ac.jp}
\orcid{0000-0002-9200-6359}
\affiliation{%
  \institution{Osaka University}
  \city{Osaka}
  \country{Japan}
}

\renewcommand{\shortauthors}{Hirota, et al.}

\begin{abstract}
Vision-and-language tasks have increasingly drawn more attention as a means to evaluate human-like reasoning in machine learning models.
A popular task in the field is visual question answering (VQA), which aims to answer questions about images. However, VQA models have been shown to exploit language bias by learning the statistical correlations between questions and answers without looking into the image content: \eg questions about the color of a \textit{banana} are answered with \textit{yellow}, even if the banana in the image is green. If societal bias (\eg sexism, racism, ableism, etc.) is present in the training data, this problem may be causing VQA models to learn harmful stereotypes.
For this reason, we investigate gender and racial bias in five VQA datasets. In our analysis, we find that the distribution of answers is highly different between questions about women and men, as well as the existence of detrimental gender-stereotypical samples. Likewise, we identify that specific race-related attributes are underrepresented, whereas potentially discriminatory samples appear in the analyzed datasets.
Our findings suggest that there are dangers associated to using VQA datasets without considering and dealing with the potentially harmful stereotypes. We conclude the paper by proposing solutions to alleviate the problem \textit{before}, \textit{during}, and \textit{after} the dataset collection process.
\end{abstract}




\maketitle

\section{Introduction}
\label{sec:intro}

\begin{figure*}[t]
  \centering
   \includegraphics[width=0.75\linewidth]{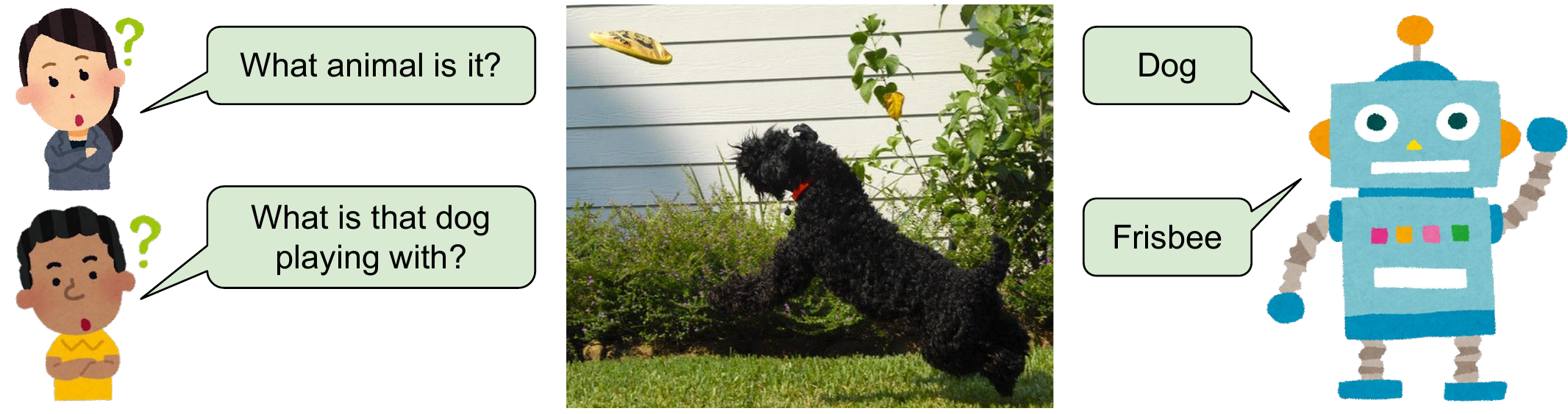}
  \caption{Diagram of VQA. Given an image and a question about the image, a model answers the question.}
   \label{fig:vqa}
\end{figure*}

The so-called vision-and-language tasks, which consist of applications that interact, process, and make decisions based on both visual and language content, present one of the greatest challenges in modern machine learning research. By construction, such applications not only need to deal with the challenges of image and text understanding but also overcome the modality gap between the visual and the language inputs. Such modality gap is non-trivial and has made tasks such as image captioning \cite{you2016image,vinyals2016show} or visual question answering \cite{antol2015vqa,malinowski2014multi} considerably popular within the computer vision (CV) \cite{wk2021clvl,wk2021vqa2vln,wk2021vqa,wk20213d} and the natural language processing (NLP) \cite{wk2021lantern,wk2021mm} communities. From classic CNN-based models \cite{vinyals2016show} to the current self-attention multi-modal Transformers \cite{lu2019vilbert,li2019visualbert}, the rapid progression of the field has only been possible thanks to the collection, annotation, and public distribution of datasets and benchmarks \cite{lin2014microsoft,chen2015microsoft,krishna2016visual,wk2021vqa,sharma2018conceptual} designed specifically to train and evaluate vision-and-language models.

The increased complexity of those models, which contain a high number of parameters to be trained, has made the availability of data a precious resource. At the same time, with the adoption of some of these models into real-world products, how this data represents the real world is of raising concern. For example, when minoritized groups are underrepresented in machine learning datasets, trained models can contribute to perpetuating social discrimination by producing potentially harmful outcomes. This has been the case for face recognition \cite{buolamwini2018gender} as well as for object recognition \cite{zhao2017men}. With respect to vision-and-language, Burns \etal \cite{burns2018women} and Zhao \etal \cite{zhao2021captionbias} showed that image captioning models, \ie models that aim to produce a descriptive sentence of a given image, perpetuate gender and racial bias. As crucial as this is, the depth of the problem has not been fully explored, and the question about whether other tasks within vision-and-language are also affected by unfair data representations remains open.

With the aim of raising awareness and starting an open discussion, this paper investigates societal bias in visual question answering (VQA), another of the foundational tasks within the vision-and-language community. The aim of VQA is to correctly answer questions about a given image (Figure~\ref{fig:vqa}), requiring understanding and associating the question and the image with potential candidate answers. One of the current challenges in VQA is that models tend to suffer from language bias~\cite{agrawal2018don}, \ie they exploit the superficial correlations between the questions and the answers in the training set, and produce answers without looking into the visual content. For example, in the VQA 1.0 dataset \cite{antol2015vqa} questions starting with \textit{what sport} are answered \textit{tennis} on the $41$\% of the samples, which leads to models learning a shortcut to always produce \textit{tennis} for this type of questions. This suggests that if societal bias, such as racism or sexism, is present in the training data, it will be highly likely perpetuated by VQA models.  Moreover, as most of the VQA datasets are collected by crowdsourcing by showing human annotators a photo and asking them to freely write questions and answers about it, it is reasonable to think that the biases from the annotators may be leaked into the data.

Specifically, this paper analyzes gender and racial bias on five VQA datasets \cite{goyal2017making,hudson2019gqa,krishna2016visual,zhu2016visual7w,marino2019ok}. Our study is based on the compilation of statistics about the representation of different demographic groups, the correlation between demographic groups and answer and question distributions, and the exploration of harmful examples within each dataset. In Section~\ref{sec:gender}, we analyze gender bias in terms of men and women representation,\footnote{We realize that gender categories should be inclusive and based on the self-identity of gender. Since VQA datasets deal with gender in binary categories and according to other computer vision studies about ethics \cite{burns2018women,zhao2017men,wang2019balanced}, we consider binary gender in this analysis.} uncovering: 
\begin{itemize}
    \item There is a systematic imbalance on gender representation. Questions about men are about twice as frequent as those about women
    in all the analyzed datasets.     
    \item Answer distributions are different between women and men questions. For example, in VQA 2.0 \cite{goyal2017making}, the answer \textit{skateboarding} appears $835$ times for questions about men, but only $60$ times for questions about women. Furthermore, we found that specific answers are co-related with each gender, \eg the percentage of answers that are \textit{blonde} is about $4$ times higher in questions about women than in questions about men.
    \item We found multiple samples that reflect traditional gender stereotypes both in men and women images.
    \item Answers based on gender stereotypes tend to appear more frequently in samples where the answers cannot be grounded in the image content, such as \textit{What is the woman thinking?}.
\end{itemize}

In Section~\ref{sec:race}, we analyze VQA datasets in terms of racial bias. 
Specifically, we dig into samples that explicitly mentions race or ethnicity. Our analysis shows: 
\begin{itemize}
    %\item Except GQA \cite{hudson2019gqa}, all the VQA datasets contain questions related to race. GQA is the only dataset whose questions and answers were automatically generated from the images, instead of written by human annotators.
   \item All the VQA datasets contain questions related to race. However, the ratio of such questions is very small for GQA \cite{hudson2019gqa}, whose questions and answers were automatically generated from the images, and for Visual7W \cite{zhu2016visual7w}, in which inappropriate samples were filtered.
   \item There is an imbalance on the representation of different demographic groups. For instance, in VQA 2.0, the answers \textit{White} or \textit{Caucasian} appear about $3.45$ times more frequently than the answers \textit{Black} or \textit{African}. 
   \item The correlation between race and nationality shows a US-centric viewpoint, with $65$\% of Black people being associated with American nationality in VQA 2.0. 
    \item We found potential harmful examples related to race, ethnicity, or nationality. As in the case of gender, these samples tend to appear when answers cannot be grounded in the visual content of the image, leading  annotators to answer the questions based on their own preconceptions of the world. 
\end{itemize}

In Section~\ref{sec:solutions}, we identify two main sources of societal bias in VQA datasets, and propose mitigation strategies to alleviate them. The first problem is the underrepresentation of minoritized groups. 
%
% introduce two possible solutions to the problem of societal bias in VQA datasets: underrepresentation of minoritized attributes, and the use of tools for detecting harmful samples. 
%As for the former, we propose to encourage VQA researchers to design models and training paradigms to avoid ignoring minoritized attributes and shortcut learning.
%As for the former, 
We emphasize the importance of recognizing societal bias in VQA datasets and taking measures against it, as models trained on such datasets can learn to ignore minoritized attributes and lead to shortcut learning \cite{geirhos2020shortcut,kervadec2021roses}.  
The second problem is the presence of harmful samples in the datasets.
% Regarding the latter, 
We propose three simple tools to remove them while considering annotation costs: 1) automatically screening to identify unanswerable questions, 2) including ethical instruction in the annotation process, and 3) creating an open platform to allow user's feedback, so that problematic samples can be easily addressed.

Finally, we would like to note that datasets are a crucial part for the development of the field, and all existing VQA datasets have been and will be necessary and important. This paper does not aim to diminish the contribution of such datasets, but to raise awareness within their users and potential dataset developers so that mitigation measures can be taken in the future.

\section{Background}

\textbf{Visual question answering (VQA)}
VQA is the task of answering a question about an image's visual content, which has been commonly used to evaluate the ability of a model to understand and integrate visual and language information. 
In the seminal work by Agrawal \etal \cite{antol2015vqa}, the first large-scale dataset for VQA was created, commonly known as VQA 1.0 dataset. VQA 1.0 contained challenging reasoning questions involving a diverse set of skills for the models to solve, such as object and activity recognition, counting, or space localization, among others.
Since models became progressively better at VQA 1.0, new datasets and challenges were proposed \cite{goyal2017making,krishna2016visual,hudson2019gqa,zhu2016visual7w,gurari2018vizwiz,marino2019ok,garcia2020knowit}, including VQA 2.0 \cite{goyal2017making}, which partly addressed the existent language bias in VQA 1.0 by increasing the diversity of answers in similar types of questions, GQA \cite{hudson2019gqa}, in which questions required various reasoning skills (\eg spatial reasoning, logical inference) to find the correct answer, and OK-VQA \cite{marino2019ok}, in which models needed to access external knowledge such as Wikipedia. 
As a result, the field has attracted a lot of attention from researches all over the world, with a large number of models been proposed ever since \cite{antol2015vqa,ben2017mutan,anderson2018bottom,kim2018bilinear,cadene2019murel,li2019visualbert,lu2019vilbert,tan2019lxmert,li2020oscar,chen2020uniter,jiang2020defense,wang2020visual,zhang2021vinvl,huang2021seeing,yang2020bert,hirota2021visual}. 

Although most of the effort has been focused on increasing the overall accuracy on the publicly available benchmarks, specially on the VQA 2.0 dataset, an important body of work \cite{agrawal2018don,clark2019don,cadene2019rubi,chen2020counterfactual,niu2021counterfactual} has been devoted to address \textit{language bias}, which refers to the existence of skewed distributions of answers with respect to a certain type of question. %(\eg the answers for the questions starting with \textit{what sport} are mostly \textit{tennis}). 
Even though VQA is a multi-modal task, because of the language bias, models tend to learn and make inferences based on superficial correlations, such as questions about \textit{bananas} are answered \textit{yellow} with high probability \cite{agrawal2018don}.  This makes the models to ignore the visual information, which prevents generalization to out-of-distribution settings. 
Manjunatha \etal \cite{manjunatha2019explicit} further examined this tendency by utilizing rule mining algorithms to correlate the questions, answers, and regions in which a model focuses. The results showed that predictions tend to contain miscellaneous rules; for example, when \textit{what} and \textit{brand} are in a question and there is a laptop in the image, the answer tends to be \textit{Dell}. 
It was also observed that in \textit{What is he/she doing?} type of questions,  answers for men were more diverse (\textit{skateboarding, snowboarding, surfing}), than answers for women (\textit{texting}). Although this evidence points at  skewed distributions in terms of gender, \textit{societal bias} in VQA datasets have not been explicitly studied yet.

\textbf{Societal bias in vision-and-language}
It is only in recent years that societal bias has been investigated in CV and NLP tasks  \cite{buolamwini2018gender,zhao2017men,wang2019balanced,wang2019racial,thong2021feature,jia2020mitigating,shankar2017no,bolukbasi2016man}. One significant study is Buolamwini and Gebru's work \cite{buolamwini2018gender} on commercial face recognition applications. They demonstrated that the system's performance varies depending on the gender and race of the individual, and in particular, misjudges women with darker skin.
 %Similarly, Zhao \etal \cite{zhao2017men} uncovered a societal bias in the image dataset and object recognition models. They investigated the correlation between gender and objects and showed not only that the objects appeared with a specific gender but also that the object recognition model amplified such bias in its prediction.
In vision-and-language tasks, there has been some recent advancements, especially for image captioning \cite{burns2018women,zhao2021captionbias,tang2021mitigating,hirota2022quantifying}.
In one of the first studies, Burns \etal \cite{burns2018women} showed that, instead of looking into the appearance of people, captioning models predicted gender words based on gender stereotypes in the contextual information. For example, when an image had a laptop, models generated the word \textit{man}, even when there was only a woman. 
Similarly, Zhao \etal~\cite{zhao2021captionbias} studied racial and gender bias. They found imbalances in gender and race representation in the COCO dataset \cite{lin2014microsoft}, with more than twice images of men than of women and $9.2$x more images of lighter-skinned than darker-skinned people.
More recently, Hirota \etal~\cite{hirota2022quantifying} proposed a metric to quantify gender and racial bias amplification of image captioning models.
%Also, they identified that models trained on this dataset amplified racial bias.
% Against this background, researches have recently been proposed to mitigate the social bias in datasets and models \cite{zhao2017men,wang2019balanced,wang2019racial,thong2021feature,jia2020mitigating}.
Given this context, in order to raise awareness and mitigate the damage societal bias may be causing to underrepresented communities, it is only natural to investigate whether other tasks and datasets within vision-and-language are also affected by this problem.

\section{Preliminaries}
% In this paper, we analyze societal bias in VQA datasets in terms of gender and race. 
We study gender and racial bias in VQA. 
We first describe the datasets under analysis in Section \ref{sec:datasets}, and then, the methodology we followed in our experiments in Section \ref{sec:methodology}.

\begin{table*}[t]
  \caption{Comparison of datasets under analysis. ${}^\dag$ denotes the images are from the intersection of COCO \cite{lin2014microsoft} and YFCC100M \cite{thomee2016yfcc100m} datasets. The types of answers can be: Open ended, which means the answers are written freely without vocabulary restrictions; Multiple choice, which means several candidate answers, with only one correct answer, are provided per question; or Closed vocabulary, which means the words used as answers are taken from a limited list.}
  \label{tab:datasets}
  \begin{tabular}{llrlrll}
    \toprule
    Year &  Dataset & Num. Images & Source & Num. QA & QA Annotation & Answer Type\\
    \midrule
    2016 & Visual Genome \cite{krishna2016visual} & $108$k & COCO${}^\dag$  & $1.7$M  & Crowdsourcing & Open ended \\ 
    2016 & Visual7W \cite{zhu2016visual7w} & $47$k & COCO & $327$k  & Crowdsourcing & Multiple choice\\
    2017 & VQA 2.0 \cite{goyal2017making}  & $204$k & COCO & $1.1$M  & Crowdsourcing & Open ended\\
    2019 & GQA \cite{hudson2019gqa} & $113$k & COCO${}^\dag$ & $22.7$M  & Automatic & Closed vocabulary\\ 
    2019 & OK-VQA \cite{marino2019ok} & $14$k & COCO & $14$k  & Crowdsourcing & Open ended \\
  \bottomrule
\end{tabular}
\end{table*}

\subsection{VQA datasets}
\label{sec:datasets}
We analyze five standard datasets, summarized in Table~\ref{tab:datasets}. Each dataset varies in the number of images, the number of questions, the annotation method, and the format of the answers; however, the images are from a common source, the COCO dataset \cite{lin2014microsoft}. A detailed description for each dataset is provided below.

\textbf{Visual Genome \cite{krishna2016visual}}
Visual Genome contains $108$k images from the intersection of COCO \cite{lin2014microsoft} and YFCC100M \cite{thomee2016yfcc100m} datasets, and $1.7$ million question-answer pairs about the images. The answers are open-ended, which means they are freely written and their vocabulary is not restricted. Questions and answers were created by human annotators following three rules: 1) questions had to start with one of the six Ws: Who, Where, What, When, Why, and How, 2) ambiguous and speculative questions had to be avoided, and 3) questions had to be precise, unique, and relatable to the image, such that they had to be answerable if and only if the image was shown.

\textbf{Visual7W \cite{zhu2016visual7w}}
Visual7W is composed of $327$k QA pairs and $1.3$M human-generated multiple choice answers on top of $47$k COCO images. The dataset is characterized by object-level rationales and multiple choice answers, which means several candidate answers are provided per question with only one being the correct one. Each question starts with one of the seven Ws: What, Where, When, Who, Why, How, and Which. Annotators instructed to create question-answer pairs while being concise and unambiguous to avoid wordy or speculative questions. After that, other annotators check the question-answer pairs to see if an average person can answer them. 

\textbf{VQA 2.0 \cite{goyal2017making}}
The dataset is built on COCO images and contains  $1.1$M question-answer pairs. The questions are categorized by question types defined by the first few words of questions (\eg \textit{What is this}, \textit{How many}). The dataset is divided into training ($80$k images and $444$k questions), validation ($40$k images and $214$k questions), and test ($80$k images and $448$k questions) sets. VQA 2.0 is the de facto benchmark for natural image VQA. When making the questions, annotators freely create questions that people can answer while making questions not easy. Also, annotators are instructed to ask questions that require the image to answer. After that, ten different annotators answer each question. The answers of the test set are not published, so we use training and validation sets in our analysis.

\textbf{GQA \cite{hudson2019gqa}}
GQA is a large-scale VQA dataset with $113$k images from the Visual Genome dataset and $22.7$M question-answer pairs. Questions require many types of reasoning which measure \eg logical inference. The question-answer pairs are automatically generated using question templates and scene graph representing all the objects and relationships in the image. Hence, the answers are limited to the words in the scene graphs, which we called closed vocabulary. Due to its large-scale, we use a random subset of roughly $20$ percent of the samples in our analysis.

\textbf{OK-VQA \cite{marino2019ok}}
The dataset is built on a part of COCO images and contains $14$k open-ended questions. Also, there are $10$ knowledge categories into which each question is classified (\eg Science and Technology, Cooking and Food). 
The dataset comprises questions that require external knowledge (\eg Wikipedia) to answer. Questions are written by human annotators following the same instructions as VQA 2.0, but the annotators are also encouraged to make questions that require external knowledge. Later, five different annotators write the answers to those questions.

\subsection{Methodology}
\label{sec:methodology}

\begin{figure*}[t]
  \centering
  \includegraphics[width=0.8\linewidth]{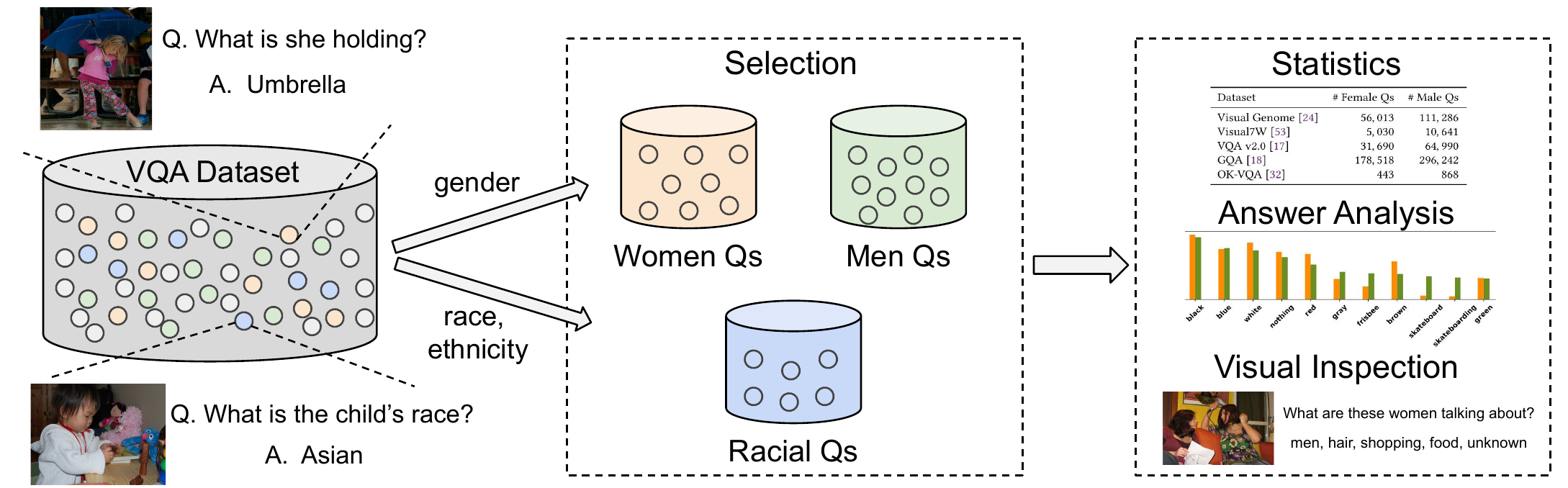}
  \caption{Overview of our methodology. For each VQA dataset, we automatically select the samples containing references to gender and race/ethnicity, and proceed to perform statistical analysis, answer analysis, and visual inspection. 
  \textcolor{DarkOrange}{Orange} and \textcolor{DarkGreen}{green} circles represent samples with a gender reference. \textcolor{SteelBlue}{Blue} circles are samples with a reference to race or ethnicity. \textcolor{DimGray}{Gray} circles are other samples.}
  \label{fig:overview}
\end{figure*}

In Figure~\ref{fig:overview}, we provide an overview of the methodology used to investigate gender and racial bias in VQA datasets. The first step is to select samples for each dataset with an explicit mention to gender (women/men questions) or race and ethnicity (racial questions). We use a rule-based approach to detect such samples. Then, we analyze gender/racial bias by comparing different sets. Concretely, we run statistics on the number of samples, analyze trends in the answers, and visually inspect for potential harmful data.

\section{Gender Bias in VQA}
\label{sec:gender}

We first analyze gender bias in VQA datasets. Following previous work \cite{burns2018women,zhao2021captionbias}, we use a  binary classification of gender, with the two gender categories being \textit{women} and \textit{men}. %\footnote{We realize that gender categories should be inclusive and based on the self-identity of gender. Since VQA datasets deal with gender in binary categories and according to other computer vision studies about ethics \cite{burns2018women,zhao2017men,wang2019balanced}, we consider binary gender in this analysis.}.
With a rule-based approach, we classify samples into women or men categories based on the gender words in the questions (if any).
Specifically, we first define a list of words for women and men, \eg \textit{woman}, \textit{girl}, \textit{she} for women, \textit{man}, \textit{boy}, \textit{he} for men.\footnote{The list of women/men words can be found in the appendix.} When a question only includes \textit{women words}, the question is classified as a \textit{women question}, and vice versa. Questions that are not classified either as women or men are excluded. Note that this analysis is based on the VQA annotator's perceived gender, and not on gender identity. We report our findings below.

\begin{table*}[t]
  \caption{Statistics of women/men questions (Women Qs/Men Qs) in VQA datasets. MoW is the number of men questions over the number of women questions. Ratio is the number of gender questions (Num. Women Qs $+$ Num. Men Qs) over the total number of questions (Num. Total Qs).}
  \label{tab:gender-ratio}
  \begin{tabular}{lrrrrrr}
    \toprule
    Dataset &  Num. Men Qs & Num. Women Qs & MoW  & Num. Gender Qs & Num. Total Qs & Ratio (\%)\\
    \midrule
    Visual Genome \cite{krishna2016visual} &  $111,286$ & $56,013$ & $2.0$ & $167,299$ & $1.7$M & $9.8$\\
    Visual7W \cite{zhu2016visual7w} & $10,641$ & $5,030$ & $2.1$ & $15,671$ & $327$K & $4.8$\\
    VQA 2.0 \cite{goyal2017making} & $64,990$ & $31,690$ & $2.1$ & $96,680$ & $658$K & $14.7$\\
    GQA \cite{hudson2019gqa} & $296,242$ & $178,518$ & $1.7$ & $474,760$ & $4.3$M & $11.0$\\ 
    OK-VQA \cite{marino2019ok} & $868$ & $443$ & $2.0$ & $1,311$ & $14$K & $9.4$\\
  \bottomrule
\end{tabular}
\end{table*}

\subsection{Questions about men are dominant}
The statistics of the number of women and men questions for each dataset are reported in Table~\ref{tab:gender-ratio}. The number of questions about men is about twice as large as the number of questions about women in all the datasets. 
This tendency is consistent with the result in \cite{zhao2021captionbias}, which shows that there are more than twice as many men images as women images in the COCO dataset \cite{lin2014microsoft}. As all the datasets in Table~\ref{tab:gender-ratio} are based on COCO images, the root of the underrepresentation of women in VQA datasets may come from the original selection of images. 
%Thus, similarly to COCO, women are underrepresented in the VQA datasets, which may lead to a marginalization of women in models trained on the datasets.
% The unfair underrepresentation of women may cause marginalization of women in models trained on the datasets because such models could learn to ignore minoritized samples.

\subsection{Answer distributions are skewed toward each gender}

\begin{figure*}[t]
  \centering
  \includegraphics[width=\linewidth]{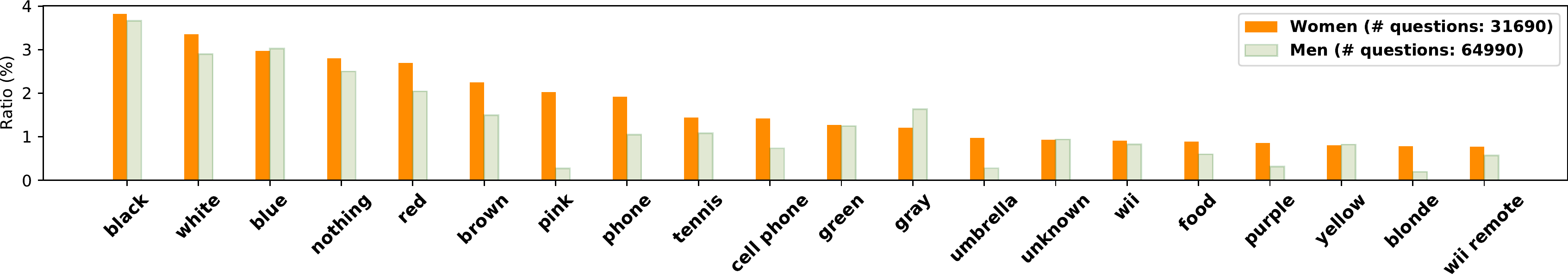}
   \includegraphics[width=\linewidth]{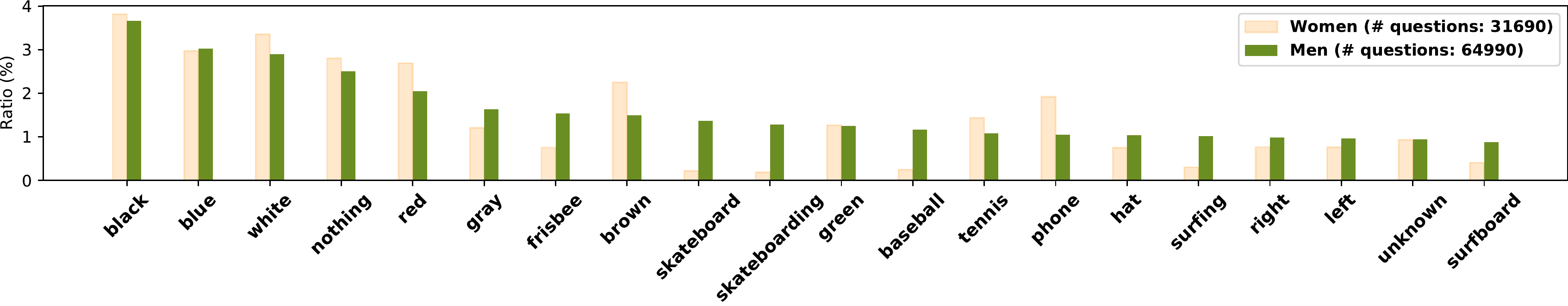}
  \caption{Top-20 frequent answers in VQA 2.0. \textbf{Above}: Frequent answers for women questions (\textcolor{DarkOrange}{orange}). For the comparison, we also show the ratio of the answers over men questions (\textcolor{DarkGreen}{green}). \textbf{Below}: Frequent answers for men questions. As in above, we also show the ratio of answers for women questions. A large difference in ratio indicates that the answer is skewed toward certain gender.}
   \label{fig:top20}
\end{figure*}

In Figure~\ref{fig:top20}, we show the top-$20$ answers for women (above) and men (below) questions in the VQA 2.0 dataset. We filter out yes/no and numeric answers. Each distribution is normalized by the number of questions for the corresponding gender. Comparing the two answer distributions, we can see that there are more frequent answers about sports in men questions (\textit{frisbee}, \textit{skateboard}, \textit{skateboarding}, \textit{baseball}, \textit{tennis}, \textit{surfing}, and \textit{surfboard}) than in women questions (\textit{tennis}). Also, the differences in the ratios about sport answers are large between the two genders, perpetuating the stereotype that sport is an activity predominantly masculine. 
The top-$20$ answers for women questions with a notably higher ratio than men are \textit{pink}, \textit{purple}, \textit{blonde}, and \textit{umbrella}, most of them strongly associated with the traditional gender stereotype of feminine.
We find these patterns are also exhibited in other datasets, \eg Figure 1 in the appendix shows that the answers about sports are more frequent in men questions in the Visual7W dataset.

\begin{figure*}[t]
  \centering
  \includegraphics[width=\linewidth]{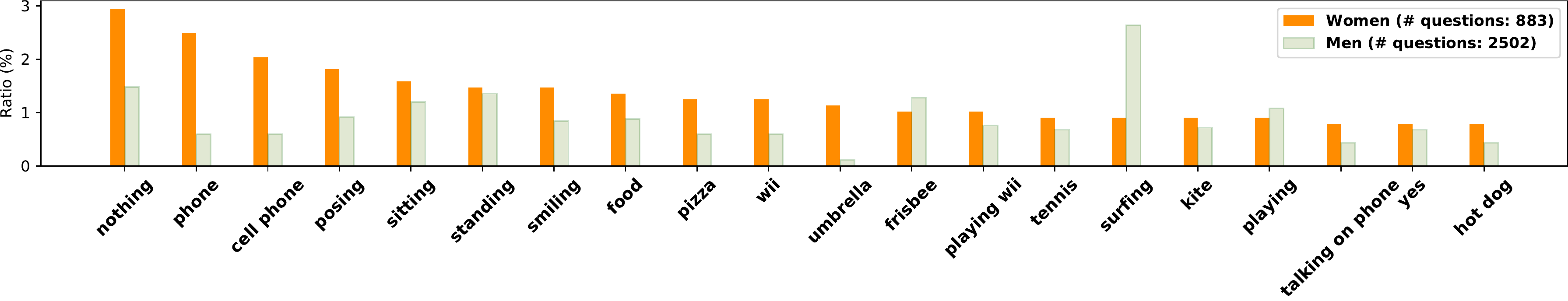}
   \includegraphics[width=\linewidth]{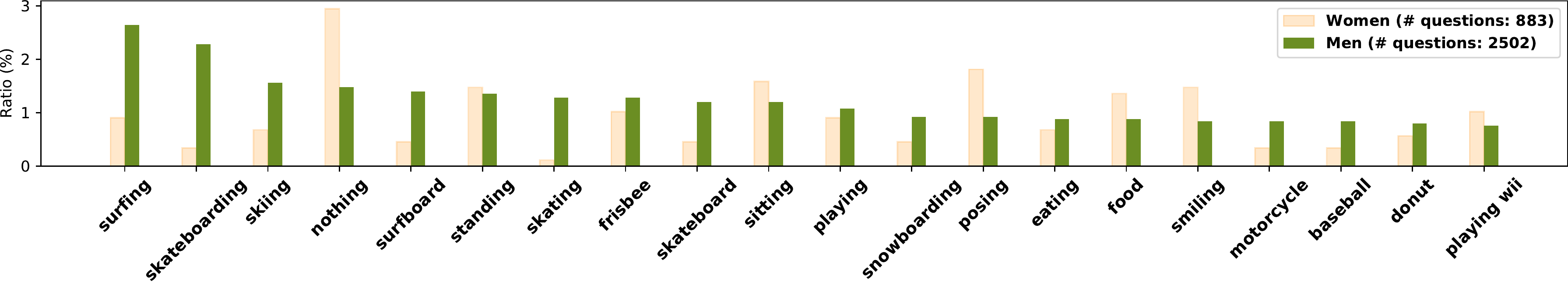}
  \caption{Top-20 frequent answers for the question type \textit{what is this} in VQA 2.0. \textbf{Above}: Frequent answers for women questions (\textcolor{DarkOrange}{orange}). \textbf{Below}: Frequent answers for men questions (\textcolor{DarkGreen}{green}). As in Figure~\ref{fig:top20}, we compare the answer ratios between the two genders.}
  \label{fig:qtype}
\end{figure*}

Figure~\ref{fig:qtype} shows the women (above) and men (below) top-20 frequent answer distribution for the specific question type  \textit{what is this} in VQA 2.0 dataset, where the skew is more prominent. Answers for men questions include multiple sports words (\eg \textit{surfing}, \textit{skateboarding}, \textit{skiing}), with higher ratios than those of women. On the other hand, in women questions more motionless words appear (\eg \textit{posing}, \textit{sitting}, \textit{smiling}, \textit{talking on phone}). % This bias may reflect the discriminatory context of modern society, where men tend to be active, and women tend to be docile.

\subsection{Gender-answer correlations reflect gender stereotypes and discrimination}

\begin{figure*}[t]
  \centering
  \includegraphics[width=\linewidth]{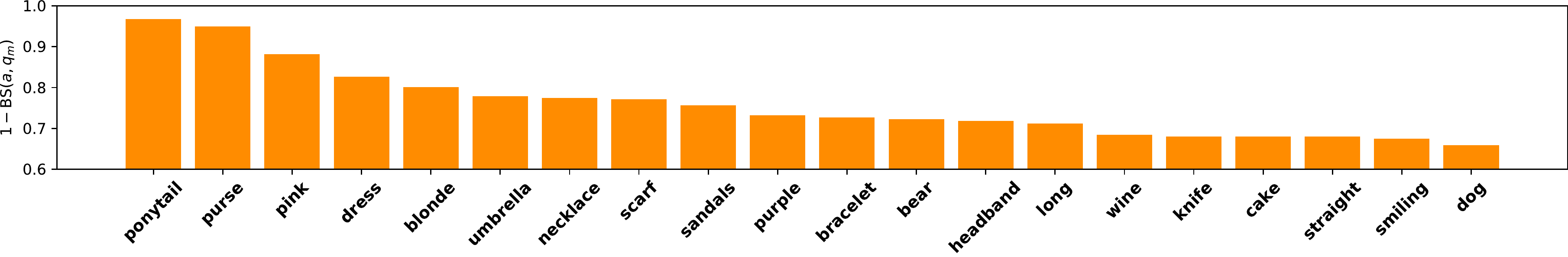}
   \includegraphics[width=\linewidth]{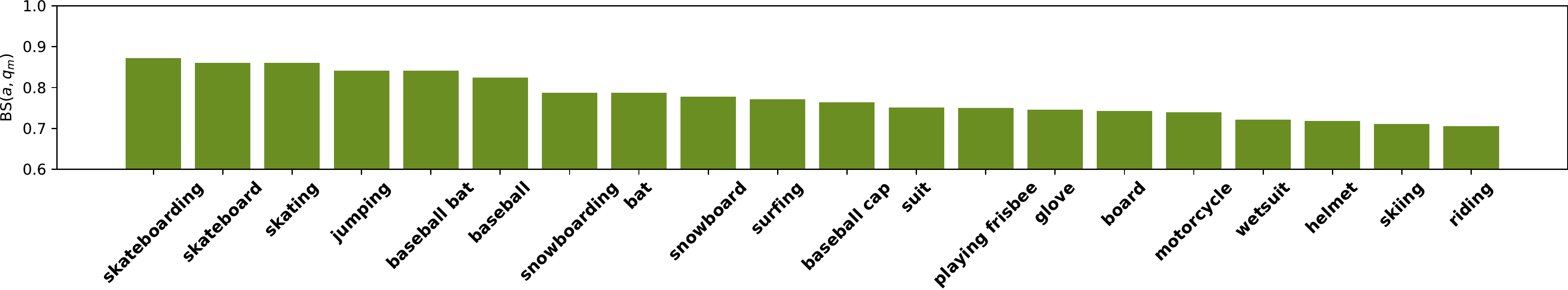}
  \caption{Top-20 answers that are co-related with women questions (\textbf{above}) and men questions (\textbf{below}) in the VQA 2.0 dataset.}
  \label{fig:ans-gender}
\end{figure*}

% This section explore answers that are correlated with each gender. 
We calculate the correlation between answers and women/men questions by utilizing the bias score (BS) defined in \cite{zhao2017men}. We adapt the definition of BS to remove the influence of the difference of the number of women/men questions. Let $q_w$ and $q_m$ denote a women question and a men question respectively, and $a \in A$ the set of the answers. We filter answers that do not appear more than $n$ times\footnote{$n$ is different among the datasets due to the size of each dataset. The detailed setting can be found in the appendix.} in women/men questions. 
Our BS gives the degree to which an answer $a$ is biased with respect to a men questions $q_m$:
\begin{equation}\label{e:shopping}
    \text{BS}(a, q_m) = \frac{c(a, q_m)}{c(a, q_m) + r c(a, q_w)},
\end{equation}
where $c(a, q)$ is the number of co-occurrences of $a$ and $q_w$ or $q_m$, and $r$ is the ratio of the number of the men and women questions (\ie $r = \text{\# men questions} / \text{\# women questions}$). If $\text{BS}(a, q_m)$ is close to $1$, then the answer $a$ is correlated with men questions. On the contrary, if $\text{BS}(a, q_m)$ is close to $0$, answer $a$ is correlated with women questions. 
The difference between BS and Figure~\ref{fig:top20} and ~\ref{fig:qtype} is that the distribution of BS shows the answers that often appear only in questions of one gender.
For example, in Figure~\ref{fig:top20}, \textit{black} is the most common answer for both gender questions, but it is not biased towards any gender, in which case the value of BS is close to $0.5$.

The top-$20$ answers that are correlated with each gender based on BS in VQA 2.0 are shown in Figure~\ref{fig:ans-gender}.
In a bias-free dataset, the distribution of BS would be flat, which reveals that
the VQA 2.0 dataset contains strong gender bias in their answers. 
Again, many sport words (\eg \textit{skateboarding}, \textit{baseball}, \textit{snowboarding}) are strongly correlated toward men questions, while no answers about sports are highly related to women questions. On the other hand, food-related words (\eg \textit{wine}, \textit{knife}, \textit{cake}) are highly correlated to women questions.
This trend is common in other datasets as well; for example, in GQA (Figure 2 in the appendix), sports or outdoor words (\eg \textit{skating}, \textit{surfing}, \textit{tan}) made up most of the top-$20$ answers for men questions, while there are no sports or activity-related answers in the case of women, and instead, static words (\eg \textit{umbrella}, \textit{sofa}, and \textit{posing}) are common. 
Similarly to the results in the previous section, these results are a reflection of the real-world stereotypes that leads to gender bias and discrimination.

\subsection{Topics of the questions are significantly different between women and men}

\begin{figure*}[t]
  \centering
  \includegraphics[scale=0.5]{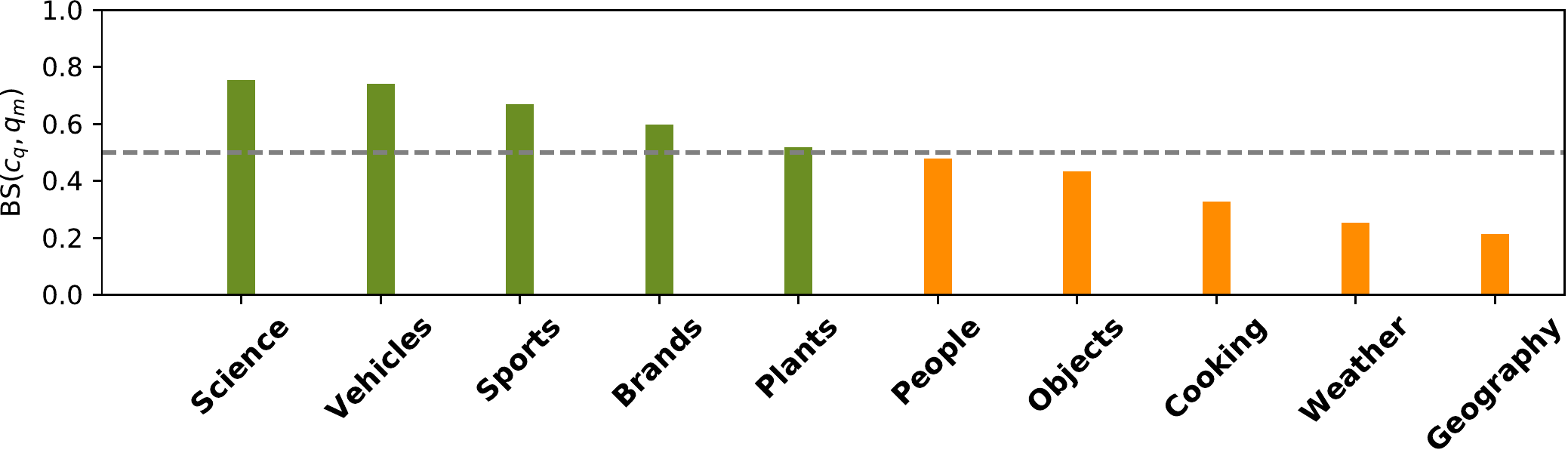}
  \caption{The degree to which each question category is biased with respect to a men questions. \textcolor{DarkGreen}{Green} color denotes the category is skewed toward men questions, while \textcolor{DarkOrange}{orange} color denotes the category is biased toward women questions. The larger the Ratio, the more the question category is biased toward men questions, and vice versa. The \textcolor{Gray}{gray} dotted line at 0.5 indicates gender balance.}
  \label{fig:ok-qtypes}
\end{figure*}

Next, we examine gender representation among different question categories. For this analysis, we use the OK-VQA dataset \cite{marino2019ok}, whose questions are categorized into 10 classes. \footnote{\textit{Science} (Science and Technology), \textit{Vehicles} (Vehicles and Transportation), \textit{Sports} (Sports and Recreation), \textit{Brands} (Brands, Companies and Products), \textit{Plants} (Plants and Animals), \textit{People} (People and Everyday life), \textit{Objects} (Objects, Material and Clothing), \textit{Cooking} (Cooking and Food), \textit{Weather} (Weather and Climate), \textit{Geography} (Geography, History, Language and Culture)} Specifically, we adapt Eq.~(\ref{e:shopping}) by replacing the answer $a$ with question category $c_q$. In other words, we calculate the degree to which each question category is biased with respect to men questions. 

Results in Figure~\ref{fig:ok-qtypes} show that some categories are strongly correlated with gender. The most correlated category with men is Science, followed by Vehicles, and Sports. On the other hand, Geography questions (which is a category gathering geography, history, language, and culture together) are biased towards women, together with Weather, and Cooking. From these observations, we conclude that men are women are differently represented in the dataset, where men are often associated with science and technology whereas women are tied to liberal arts and cooking \cite{van2016boys}.
%mens are tied to the sciences, while womens are tied to the liberal arts fields of study in this dataset. In addition, the questions about cooking and food are also biased toward womens, representing the societal bias that those who cook tend to be women.

\subsection{Gender-stereotypical samples}

Aside from the statistics in the preceding sections, we manually explore each dataset for potentially sexist or harmful samples. Specifically, we investigate more than $300$ samples for each dataset and we find that the datasets in which different annotators generate the questions and the answers separately (\ie VQA 2.0 \cite{goyal2017making} and OK-VQA \cite{marino2019ok}) often contain gender-stereotypical samples. Also, Visual Genome \cite{krishna2016visual}, which is no filtering to ensure questions are visually grounded, contains such samples. 
On the contrary, GQA \cite{hudson2019gqa}, which automatically generates samples, and Visual7W \cite{zhu2016visual7w}, which filters out not visually grounded samples, do not contain such samples.  

%\begin{figure}[t]
%  \centering
%  \includegraphics[width=\linewidth]{figures/VQAv2/vqav2-gender-stereo (7).pdf}
%  \caption{Examples of gender stereotypes in VQA 2.0 \cite{goyal2017making}.}
%  \label{fig:vqa-gender-samples}
%\end{figure}

%\begin{figure}[t]
%  \centering
%  \includegraphics[width=\linewidth]{figures/OKVQA/ok-vg-gender-stereo (8).pdf}
%  \caption{Examples of gender stereotypes in OK-VQA \cite{marino2019ok} (\textbf{above}) and Visual Genome \cite{krishna2016visual} (\textbf{below}). }
%  \label{fig:ok-gender-samples}
%\end{figure}

\begin{figure*}[t]
  \centering
  \includegraphics[width=\linewidth]{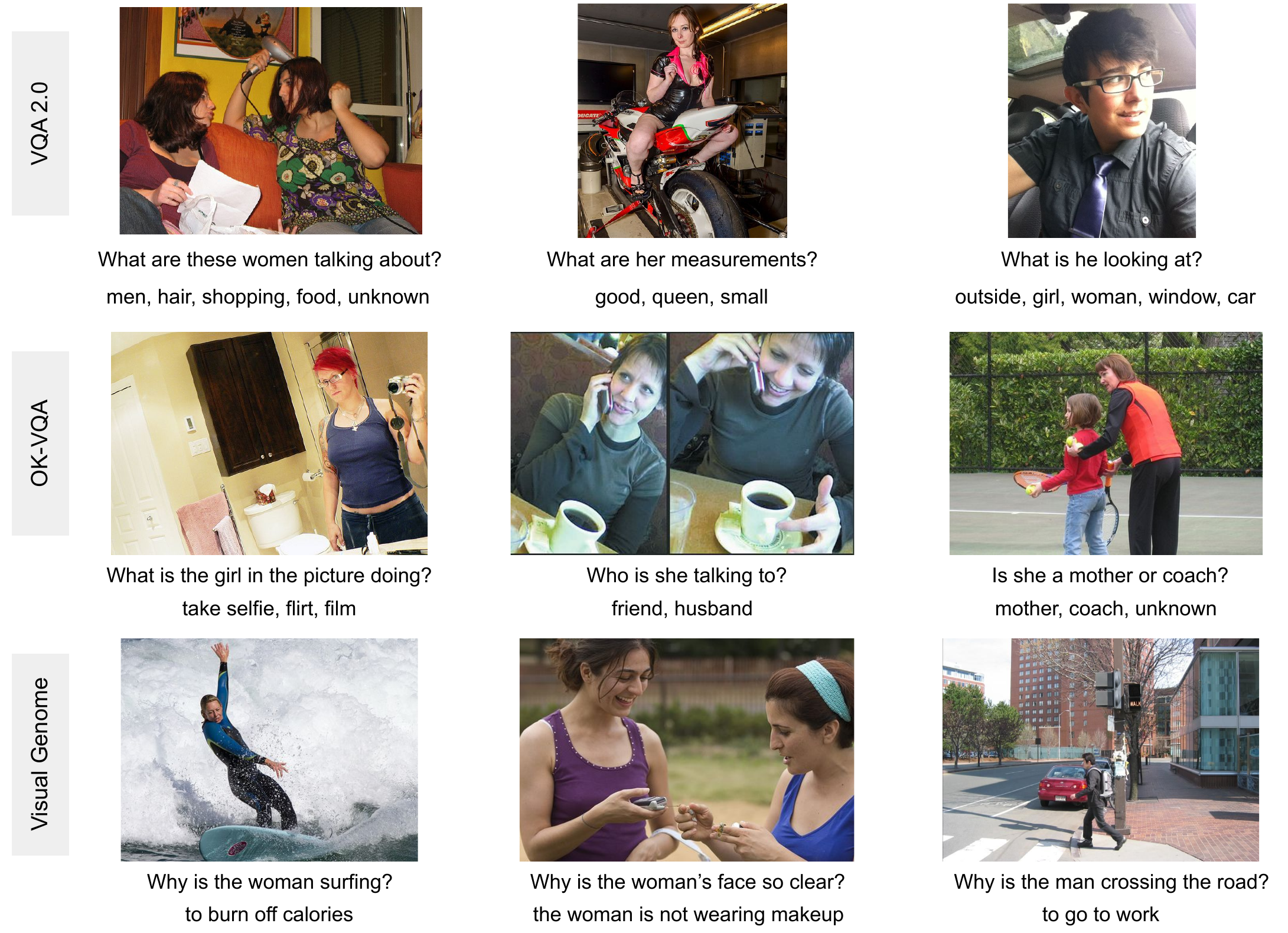}
  \caption{Examples of gender stereotypes in VQA 2.0 \cite{goyal2017making} (\textbf{above}), OK-VQA \cite{marino2019ok} (\textbf{middle}), and Visual Genome \cite{krishna2016visual} (\textbf{below}). }
  \label{fig:vqa-ok-gender-samples}
\end{figure*}

Some examples, consisting of an image, a question, and ground truth answers, are shown in Figure~\ref{fig:vqa-ok-gender-samples}.
More examples can be found in the appendix. In the top-row left example, 
to the question about the topic of the conversation between the two women, which cannot be inferred using visual clues, some annotators provided gender-stereotypical answers such as \textit{men}, \textit{hair}, and \textit{shopping}, probably based on their own prejudices about women. %Since there are no clues in the image to support those answers, it is hypothesized that the answers are based on the stereotypes of women held by the annotators. 
%Also, the middle right example asks who the woman in the image is, and the answers contain \textit{mom} and \textit{mother} even though we cannot see any evidence to support these answers. They may also come from the stereotypes of women by the annotators. 
Similarly, in the top-row right example of Figure~\ref{fig:vqa-ok-gender-samples}, the question asks what the man is looking at, but it cannot be seen in the image. In this case, the provided answers contain the words \textit{girl} and \textit{woman}. 
%The common denominator in all these examples is that it is impossible to answer those questions just by looking at images. The annotators may use their own stereotypes to answer such questions.
The common denominator in these examples is that their questions are not visually grounded: there is no definite and unambiguous evidence in the image to answer. 

Additionally, we find some inappropriate or directly harmful examples. For instance, the top-row middle question in Figure~\ref{fig:vqa-ok-gender-samples} has a sexual connotation about the woman in the image, and in the second-row left image in Figure~\ref{fig:vqa-ok-gender-samples}, where a woman is just standing in front of a mirror with a camera, one of the ground truth answers to the question of what is she doing is \textit{flirt}, implying that the mere fact that a woman is standing is to seduce someone.

\subsection{Discussion}

\textbf{Biased distributions }
We have shown that the ratio of men questions over women questions and the answer distributions are biased in all datasets. 
Especially for the ratio of men questions over women questions, we know the real-world ratio of the men to women is roughly $1:1$.
Hence, the skew of the ratio of men questions over women questions in VQA datasets (\ie men questions are about twice as numerous as women questions) is far from the real world distribution. That is unfair underrepresentation of women and could lead machine learning models trained on these datasets to potentially ignore women. This means that models perpetuate the underrepresentation of women unveiled in the dataset, which is very problematic from the perspective of gender equality. 

%On the other hand, we hardly know the real-world distributions in most cases. 
As for the answer distributions, while we have shown that the answer distributions are highly skewed toward each gender, it is difficult to know the real-world distributions (\eg the actual gender ratio among those who snowboard). Furthermore, realistically, the gender ratio for the answers cannot be aligned, and even if it could be, the gender bias in models might not disappear \cite{wang2019balanced,buolamwini2018gender}. 
However, machine learning models trained on these datasets without considering the biased distributions can learn to ignore underrepresented combinations in the datasets (\eg a woman who snowboards) and lead to shortcut learning \cite{geirhos2020shortcut,kervadec2021roses}. 
Being aware of such bias toward each gender encourages the community to design better model architectures and training paradigms to mitigate the bias, which is a research direction to be further explored, as in \cite{burns2018women}. 

\textbf{Gender-stereotypical examples }
Compared to the skewed answer distributions, gender-discriminatory samples are undoubtedly harmful. 
Such samples are often found when the associated question is not visually grounded. For such questions, annotators may answer based on their gender stereotypes.
% Also, we show that there are samples whose questions/answers themselves can sexualize women.
Such harmful samples are found in datasets in which not visually grounded questions are not filtered (\ie VQA 2.0, OK-VQA, and Visual Genome). % The main difference between these three datasets and the others \cite{hudson2019gqa,zhu2016visual7w} is that the latter filters out questions that cannot be answered by looking at the images (\ie Visual7W \cite{zhu2016visual7w}) or has no way to include such samples by design due to the automatic generation of question-answer pairs from scene graphs (\ie GQA \cite{hudson2019gqa}).
Also, in VQA 2.0 and OK-VQA, the annotators who create the questions are different from those who answer them. This choice is to deal with the problem of multiple possible correct answers to the same question \cite{antol2015vqa}. Although the process allows the datasets to have diverse answers, it does not require the annotators to answer to their own questions and gives room to make questions that are not visually grounded.
On the other hand, in GQA and Visual7W, while not visually grounded questions hardly exist because of the datasets construction, the diversity of answers is limited.
In conclusion, manual removal of potential harmful questions may be necessary to ensure ethical goodness. 
%In that way, VQA datasets can remove harmful gender-stereotypical samples even if the annotators who create questions are different from those who answer them.
Thus, it may be a good practice to build an efficient mechanism to report potential ethical problems and review them on a regular basis to decide whether some samples should be removed.

\section{Racial Bias in VQA}
\label{sec:race}

\begin{table*}
  \caption{Statistics of racial questions (Racial Qs) in VQA datasets. Ratio is the number of racial questions over the number of questions.}
  \label{tab:racial-questions}
  \begin{tabular}{lrrr}
    \toprule
    Dataset &  Num. Racial Qs & Num. Total Qs & Ratio (\%)\\
    \midrule
    Visual Genome \cite{krishna2016visual} & $619$ & $1.7$M  & $0.43$ \\
    Visual7W \cite{zhu2016visual7w} & $74$ & $327$K & $0.05$\\ 
    VQA 2.0 \cite{goyal2017making} & $799$ & $658$K & $0.12$\\
    GQA \cite{hudson2019gqa} & $85$ & $4.3$M & $0.00$\\ 
    OK-VQA \cite{marino2019ok} & $30$ & $14$K & $0.21$\\
  \bottomrule
\end{tabular}
\end{table*}

To study racial bias, we first identify samples in the dataset with a reference to race or ethnicity.  We select all the samples whose questions explicitly contain the words \textit{race} or \textit{ethnicity}. We refer to these samples as \textit{racial samples} or \textit{racial questions}. Additionally, we elaborate a list of \textit{racial-related words} (\eg \textit{Asian}, \textit{Caucasian}, \textit{Black}) and \textit{nationality-related words} (\eg \textit{American}, \textit{Chinese}, \textit{Indian}) from the answers in VQA 2.0.\footnote{The complete list of words can be found in the appendix.} Our findings are reported below. 

\subsection{Most datasets contain racial words}
The number of racial samples per dataset is shown in Table~\ref{tab:racial-questions}. All datasets contain racial questions. However, the ratio of racial questions is very small for GQA, whose questions and answers were created automatically from the image's scene graph, and for Visual7W, in which annotators had explicit instructions to write visually grounded questions, and inappropriate samples were filtered. In contrast, VQA 2.0, Visual Genome, and OK-VQA show higher ratios of racial questions. As the absolute number in OK-VQA is small, we conduct our analysis in the VQA 2.0 and Visual Genome.
%define racial questions asking about the race/ethnicity of people in images.
%Through the manual exploration of the questions whose answers contain racial terms (\eg African, Caucasian, Asian), we identify that those questions tend to include \textit{race} or \textit{ethnicity}. Therefore, we define racial questions as ones containing these two words. 
%We also define racial words (\eg Asian, Caucasian, Black) and nationality words (\eg American, Chinese, Indian) for the analysis in this section based on the answer list of VQA 2.0 \cite{goyal2017making}. The complete lists of racial and nationality words are in the appendix. 

\begin{figure*}[t]
  \centering
  \includegraphics[scale=0.5]{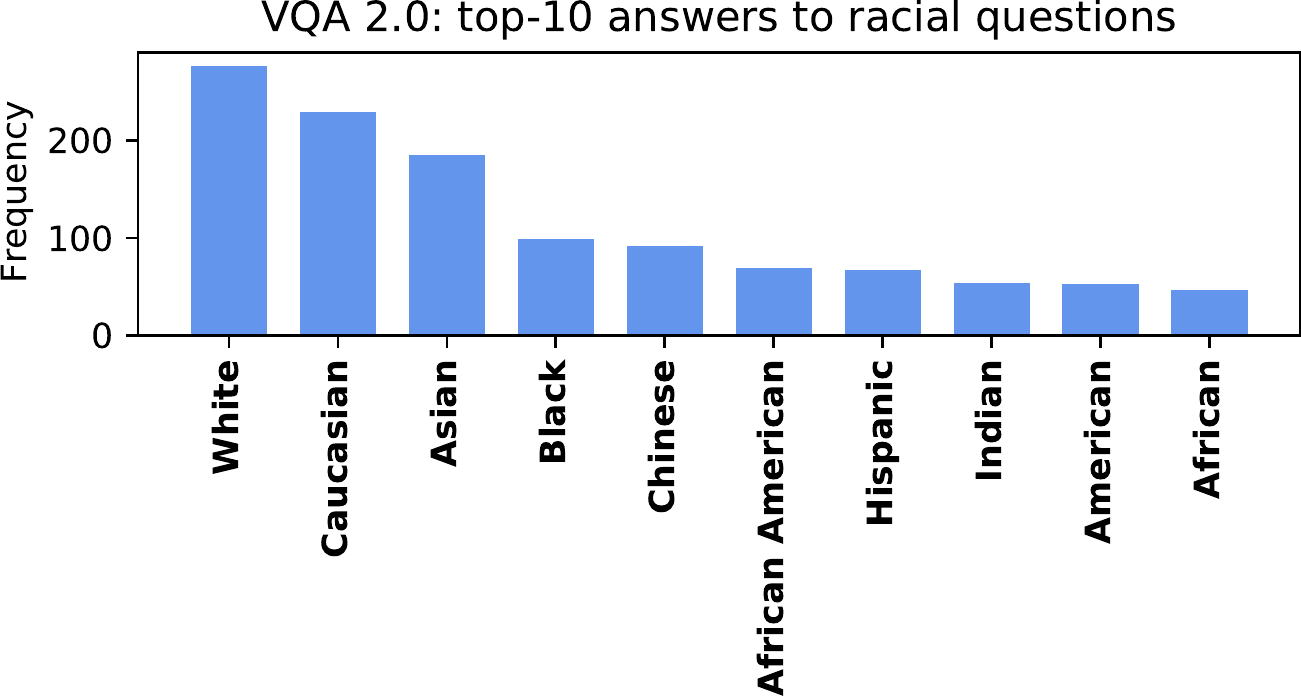}
  \hspace{10pt}
   \includegraphics[scale=0.5]{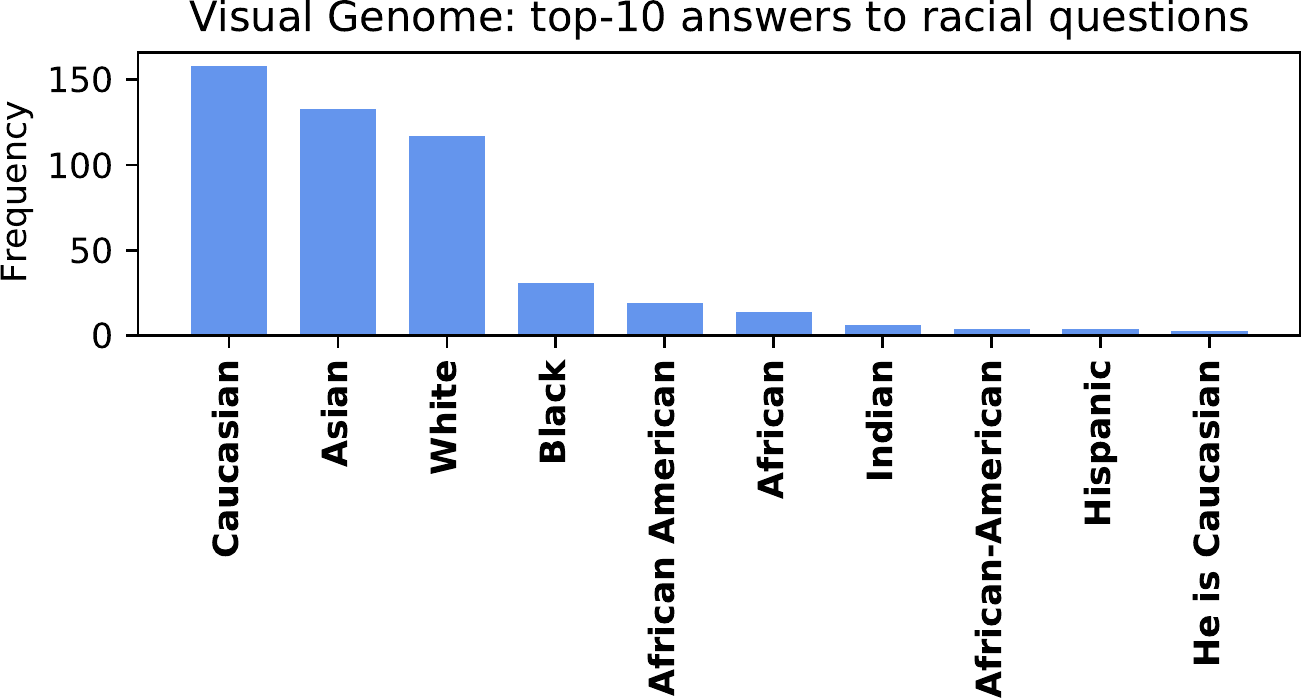}
  \caption{Top-$10$ answers to racial questions in VQA 2.0 \cite{goyal2017making} (\textbf{left}) and Visual Genome \cite{krishna2016visual} (\textbf{right}).}
  \label{fig:race-top10}
\end{figure*}

\subsection{White people are majority, and Black people are minority}

First, we investigate the distribution of answers for racial questions. To remove answers that may not be related to race, ethnicity, or nationality, we filter \textit{yes}, \textit{no}, \textit{horse}\footnote{When \textit{horse} is in the answers, the questions are more likely to ask about horse race rather than race of individuals.} and answers to questions about colors\footnote{This is to avoid confusing answers about color with race.}. In Figure~\ref{fig:race-top10}, we show the top-$10$ answers to racial questions in VQA 2.0 and Visual Genome. 
In both datasets, the demographic group that occupies the largest amount of answers is related to White people (\textit{White}, \textit{Caucasian}), followed by Asian people related words (\textit{Asian}, \textit{Chinese}). On the contrary, words usually associated with Black people (\textit{Black}, \textit{African American}, \textit{African}) and Hispanic people (\textit{Hispanic}) appear less frequently, showing an underrepresentation of darker-skinned people on the analyzed samples. This tendency has also been observed in other computer vision datasets, such as facial recognition (e.g., $79.6$\% subjects are lighter-skinned in IJB-A \cite{klare2015pushing,buolamwini2018gender}), or in image captioning \cite{zhao2021captionbias}. %As with the underrepresentation of women discussed in Section~\ref{sec:gender},
This imbalance can lead to poor performance on images of darker-skinned people in models trained on such datasets.

\subsection{US-centric perspective of nationality and race}

% By exploring the answers to racial questions, we find that when a question asks about the race or ethnicity of a person in an image, some annotators tend to answer nationality, not a race, in VQA 2.0 \cite{goyal2017making}. Therefore, 
We next explore the relationship between race and nationality in the VQA 2.0 dataset. We examine the co-occurrence of nationality-related and racial-related words in the answers. We roughly categorize them into three races or ethnicities: African-oriented (with the words \textit{Black} and \textit{African}), Asian-oriented (with the words \textit{Asian} and \textit{Oriental}), and White-oriented (with the words \textit{White} and \textit{Caucasian}).\footnote{More categories are not include due to the limited number of the samples.} % We show the top-$10$ nationality answers co-occurring with racial words in each category in Figure~\ref{fig:race-nation}. 

\begin{figure*}[t]
  \centering
  \includegraphics[width=0.87\linewidth]{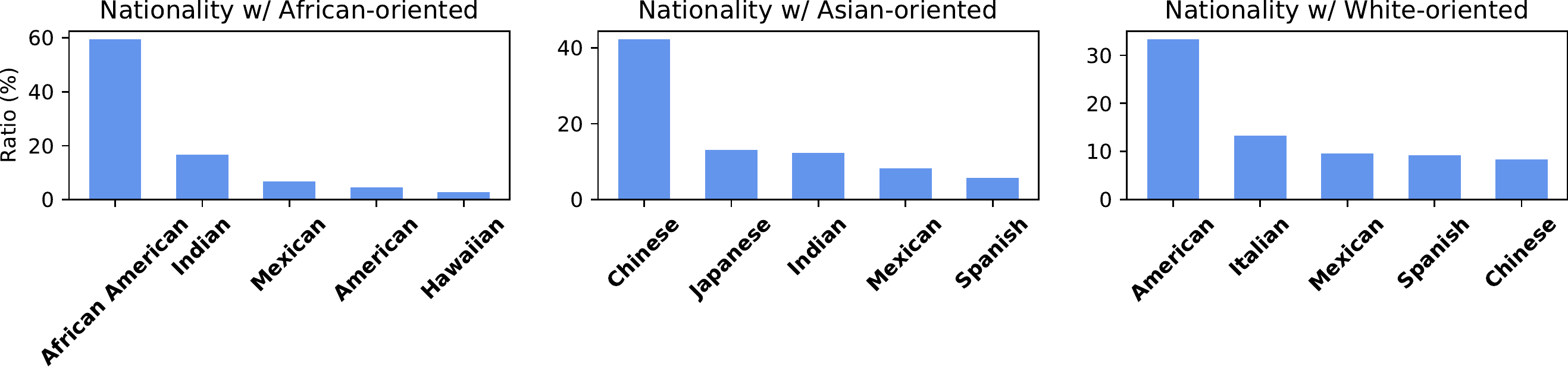}
  \caption{Top-$5$ answers related to nationality that co-occur with African-oriented (Black, African) (\textbf{left}), Asian-oriented (Asian, Oriental) (\textbf{middle}), and White-oriented (White, Caucasian) (\textbf{right}) in VQA 2.0 \cite{goyal2017making}.}
  \label{fig:race-nation}
\end{figure*}

Figure~\ref{fig:race-nation} shows the top-$10$ nationalities co-occurring with each racial category.  
Each racial category is strongly tied to a specific country. 
%For African-oriented, the most common nationality answer is African American . 
The result in Figure~\ref{fig:race-nation} (left) shows that about $65$\% of Black people are considered to be American (\ie \textit{African American} or \textit{American}). 
As for Asian-oriented category, most of the nationality answers are Chinese with a 42\% ratio, followed by Japanese and Indian with a 13\% and 12\% ratio, respectively (Figure~\ref{fig:race-nation} (middle)).
%Specifically, $42$\% of Asians are deemed as Chinese, and $13$\% and $12$\% of them are considered Japanese and Indian, respectively.
Regarding the White-oriented category, American is the most frequent nationality answer with a $33$\% ratio (Figure~\ref{fig:race-nation} (right)). %This indicates that White is highly connected to Americans. 
As the concept of race is highly tied to the social and cultural background of each individual \cite{hanna2020towards}, it is remarkable to note that the relationship between race and nationality in the analyzed datasets seems to be rooted in a United States point of view where White and Black people are associated with American nationality, and Asian people with Chinese nationality. This is probably the result of a US-centric annotation process. %, with $93$\% of the annotators in Visual Genome being from the United States \cite{krishna2016visual}.

% This shows that the relationships between race and nationality is based on a US-centric point of view. The first reason is that Black and White people are judged to be American in the dataset. The other reason is that Chinese people make up the largest percentage of Asian people living in the United States \cite{us-census}. \alert{Another important fact that follows our hypothesis is that the distribution of Visual Genome's annotators is very skewed toward Americans ($93$\% of the annotators are from the United States). } 
% The US-centric race-nationality bias come to a problem when an annotator answers nationality questions even though there is no evidence to specify the nationality. In fact, we can find such samples in several VQA datasets (\eg the middle left example in Figure~\ref{fig:race-sample}). The race-nationality questions without visual grounding can lead  models to learn the undesirable bias.

\subsection{Racial-stereotypical examples}

\begin{figure*}[t]
  \centering
  \includegraphics[width=\linewidth]{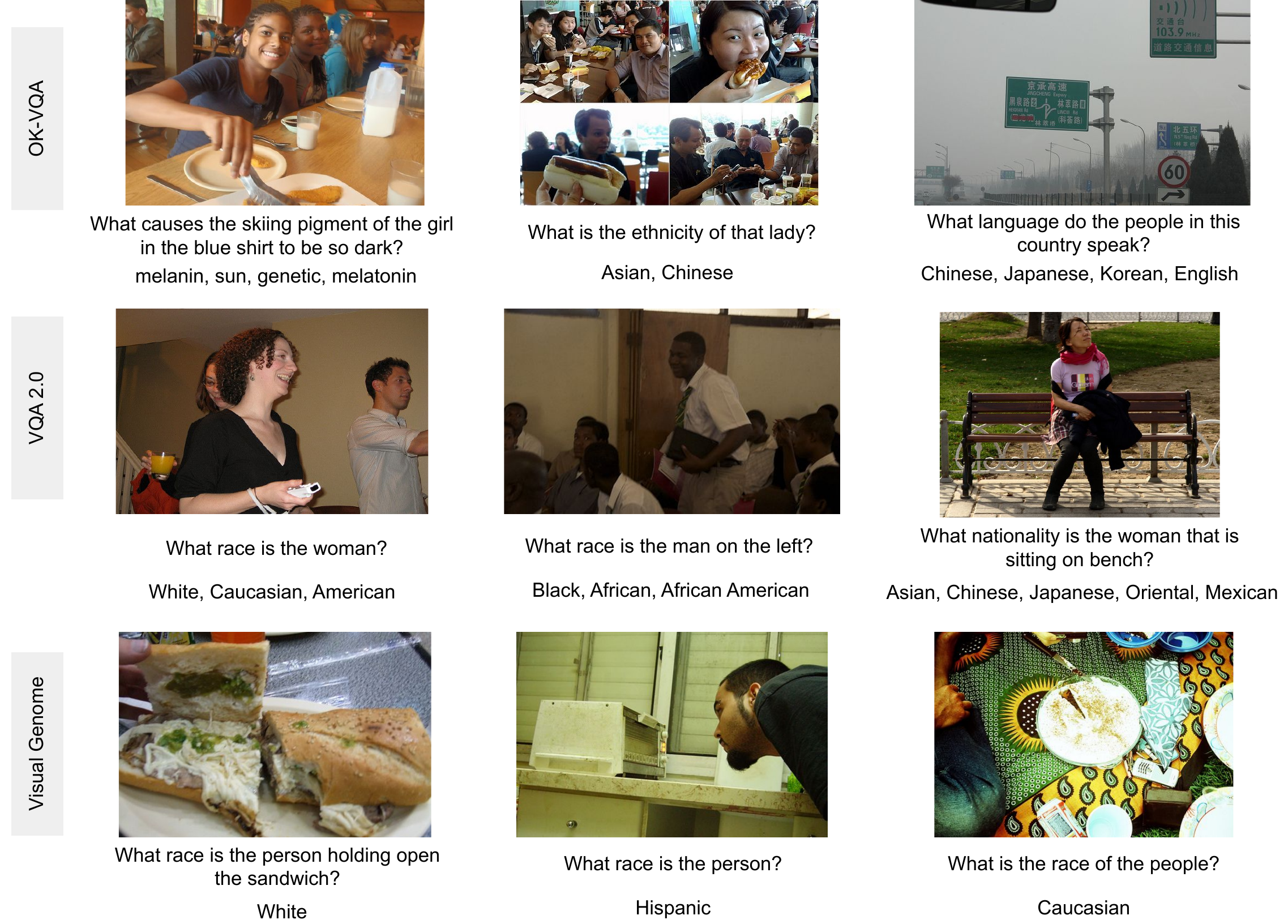}
  \caption{Racial samples in OK-VQA \cite{marino2019ok} (\textbf{above}), VQA 2.0 \cite{goyal2017making} (\textbf{middle}), and Visual Genome \cite{krishna2016visual} (\textbf{below}).}
  \label{fig:race-sample}
\end{figure*}

We manually inspect all the racial samples for all the datasets to check whether they can be potentially harmful. In addition to this, we conduct an intersectional analysis and explore more than $300$ samples of women and men questions for each dataset in terms of racial bias. We find that there are two types of samples with racial bias: 1) racial discriminatory samples, and 2) biased judgment samples. Such examples appear in the VQA 2.0, Visual Genome, and OK-VQA datasets, and some of them are shown in Figure~\ref{fig:race-sample}. More can be found in the appendix. 

Some samples that fall into the \textit{racial discriminatory} category are shown in Figure~\ref{fig:race-sample}. For example, in the top-row left example, the question \textit{What causes the skiing pigment of the girl in the blue shirt to be so dark?} implies that lighter-skin is the standard. Another example is shown in Figure~\ref{fig:race-sample} second-row right image, whose question asks about the woman's nationality. One of the answers, \textit{Oriental}, is an outdated term that has not been used in US federal laws since $2016$ \cite{oriental}. 
%is considered to be offensive, especially when used to call Asian people. The use of \textit{Oriental} has been prohibited in US federal laws \cite{oriental}, 
Still, it appears $23$ times as an answer in VQA 2.0.
With respect to \textit{biased judgment} samples, they often appear in questions that are not visually grounded. In other words, when there is no clue in the image to identify the race, ethnicity, or nationality. For example, in Figure~\ref{fig:race-sample} bottom-row left image, the answer is \textit{White} even though we can only see the tips of the fingers. A similar case can be seen in the bottom-row right example. % Although they should not answer to such a question without clear evidence, the annotators tend to answer based on their stereotypes and assumptions about race.

\subsection{Discussion}

\textbf{Biased distributions}
In VQA 2.0 and Visual Genome, the number of samples related to White people is much greater than samples related to Black and Hispanic people (\eg there are $3.45$x more samples related to White people than Black people in VQA 2.0). 
The skewed distribution of race can be problematic if models trained on those datasets are used in real world applications, as the underrepresentation of certain races or ethnicities may lead to biased answers. 
Although it seems ideal to have a uniform racial distribution, race itself is a vague concept, which is strongly tied to the personal background of each individual \cite{hanna2020towards}. Furthermore, aligning racial distributions alone is insufficient to remove racial bias from the models \cite{buolamwini2018gender,wang2019balanced}. For these reasons, it is essential to make an effort to have racial diversity in the datasets, but at the same time, it is vital to devise learning strategies that can debias the datasets.

\textbf{Biased samples}
Even though the number of racial questions is relatively small, the stereotypical samples in terms of race are unquestionably harmful and should be removed from the datasets. 
Datasets that have not gone through manual screening to remove samples that are not visually grounded are the most affected ones (VQA 2.0, Visual Genome, and OK-VQA). Also, the problem is accentuated in VQA 2.0 and OK-VQA, which increased the diversity in their answer set by making different annotators to answer the written questions. 
As samples with racial discrimination or biased judgment of race/nationality may reflect the bias in the annotators, an unconstrained or less-constrained annotation process may be prone to contain such harmful samples. This is supported by the fact that, we could not find such samples in GQA, which automatically generates question-answer pairs, or Visual7W, which applies manual filtering to exclude questions that are not based visually grounded. 
%As for biased judgment of race or nationality, the samples also exhibit annotators' bias about race. Such samples tend to happen when the questions are not visually grounded, where there is no definite and unambiguous evidence to answer in an image. 
Because of this, an additional cleansing process on the samples may be a necessary strategy to remove samples with racial discrimination and biased judgment.

\textbf{Necessity versus validity of asking questions about race}
We take a step back and cast doubt on asking about race in the first place. 
%Indeed, as long as the concept of race exists in reality, and people can recognize race to some extent (although there is a great deal of bias), it seems necessary to ask questions about race.
%Indeed, race is a recognized concept to characterize people and being proud of ourselves, so our origin is a natural attitude; therefore, it is understandable? to ask a question about one's race/ethnicity. 
% Indeed, race is a recognized concept to characterize people and being proud of ourselves, so our origin is a natural attitude. Therefore, it is understandable to ask a question about one's race/ethnicity?. 
%However, race is so ambiguous that it is nearly impossible for a person to determine race of the other individuals correctly.
Although race has been used to categorize people for a long time, it is extremely hard to provide fixed categories in which people from different backgrounds fit together \cite{khan2021one}. 
Moreover, visual information alone is hardly sufficient to identify one's race, ethnicity, or nationality, so asking this type of question is prone to cause biased answers based on stereotypes.
%In fact, as shown in Table~\ref{tab:racial-questions}, GQA \cite{hudson2019gqa} and Visual7W \cite{zhu2016visual7w}, which are less likely to include questions that cannot be answered by looking at the image, have fewer racial questions.
% In addition, due to the ambiguity of race, answers to racial questions tend to reflect the preconceived notions held by annotators.
Thus, we believe that, at least in VQA datasets, questions about race should be discouraged. %, especially when there is no pressing need to do so because it may promote racial bias. 
%If questions about race are to be included, care should be taken to implement rigorous screening. %and diversity of racial backgrounds of annotators.

\section{Possible Solutions}
\label{sec:solutions}

In order to reduce the risks associated with gender and racial bias, we would like to encourage VQA researchers to increase their awareness to this problem and take steps to address it.
We specifically discuss possible solutions to address the two major problems presented in this paper: skewed distributions and harmful samples.

\textbf{Skewed distributions}
We have shown that the analyzed VQA datasets have distributional bias related to gender (Section~\ref{sec:gender}) and, some of them, to race (Section~\ref{sec:race}).
Aligning the distributions is tricky for many reasons, such as the existence of bias in the real-world and the ambiguity of race categorization.
Furthermore, even if the distributions could be aligned, this is not sufficient for bias-free models; models can still amplify bias.
Nevertheless, if the distributions are too skewed in gender and race, models are  more likely to ignore underrepresented groups of people and increase the risk of shortcut learning.
Therefore, efforts should be made to avoid this underrepresentation. 
More importantly, we encourage the users of the datasets to be aware of the distributional biases related to gender and race in VQA datasets and design models and training paradigms that can address these issues.

\textbf{Harmful samples}
We have found that some of the VQA datasets contain harmful samples that exhibit gender or racial stereotypes. 
Such samples are often found when the associated question is unanswerable from the image content. Thus, datasets that have no filtering processes, such as VQA 2.0, Visual Genome, OK-VQA, are prone to contain such samples. 
Also, questions and answers themselves can be discriminatory.
The ideal solution is to remove harmful samples by conducting a manual filtering, but the cost for such a process can be extremely expensive, especially if the size of dataset is large. 
%
% However, making such screening mandatory and putting the onus on dataset developers to create datasets containing no harmful samples would be very costly, especially if the size of datasets is large.
% In other words, if we make it the responsibility of developers to create completely ethical datasets, researchers without large amounts of funding will not be able to create datasets, which will lead to the marginalization of such researchers.
% In addition, datasets play a central role in machine learning. If only such complete datasets are allowed to be released to the public, the development of datasets will be significantly hampered.
So, we propose three alternative solutions to address both the ethical and the cost problems: 1) \textit{automatic screening}, 2) \textit{ethical instructions}, and 3) \textit{a feedback platform for users}.

For the automatic screening, we propose to train a model to identify unanswerable questions from images. To train such a model, a labeled dataset to identify whether a question is answerable from an image might be necessary. With a trained model, visually not grounded samples could be potentially filtered out. Although the model's performance is not guaranteed, it could be used to ease the manual screening process as a pre-filtering step.

The second proposed solution is to incorporate ethical instruction in the dataset's annotation process. Ethical instructions are not commonly provided to annotators when creating VQA datasets. Nevertheless, as we have shown, VQA datasets can contain harmful samples; so instruction for annotators to be aware of making ethical questions and answers could reduce the amount of harmful samples. 
%Some examples of the ethical instructions include:
%\begin{itemize}
%    \item To not ask questions and answers that you would not if a person in an image was of a different gender or race.
%    \item To not speculate an answer if you do not recognize an answer to a question about people in an image.
%    \item After thinking of questions or answers, to make sure they do not reflect gender or racial stereotypes.
%\end{itemize}
In this paper, we only focused on gender and race as demographic attributes, but ethical instructions should be extended to make datasets fairer with respect to any other attributes.

The last solution is to create a platform to report potential ethical problems and review them to decide whether they should be removed. 
The platform should allow dataset users to report and share with dataset developers when they find harmful samples in their use or investigation of datasets.
This platform would be based on the idea of shifting from the traditional developer-driven paradigm of dataset creation to a user-participatory paradigm.
Incorporating a process that allows users to improve datasets can solve both cost and ethics issues at a high level.
% Thus, the platform will allow the creation of datasets without marginalizing any but the most well-funded researchers.

\section{Conclusion}
\label{sec:conclusion}

We investigated gender and racial bias in VQA datasets through the compilation of statistics and the manual exploration of harmful samples. 
The results showed: 1) distributions are very skewed concerning gender or race, and 2) harmful samples, denoting annotators' gender or racial stereotypes, exist in VQA datasets. 
Additionally, we discussed potential solutions. 
We proposed the automatic screening of samples, the inclusion of ethical instructions in the annotations process, and the creation of a platform for receiving user's feedback.
Through the analysis and discussion in this paper, we hope to raise awareness and encourage the VQA research community to take measures to mitigate societal bias.

%\vspace{-14pt}
\paragraph{Funding}
This research was partially supported by JST CREST Grant No.~JPMJCR20D3 (JST) and JSPS KAKENHI No.~JP22K12091 (JSPS). The JST and JSPS had no role in the design and conduct of the study; access and collection of data; analysis and interpretation of data; preparation, review, or approval of the manuscript; or the decision to submit the manuscript for publication. The authors declare no other financial interests.

\bibliographystyle{ACM-Reference-Format}
\bibliography{sample-base}

\appendix

\section{Gender Bias in VQA}

\paragraph{Gender words}
We list the female/male words that are used to define female/male questions:  
\textcolor{orange}{woman}, \textcolor{orange}{female}, \textcolor{orange}{lady}, \textcolor{orange}{mother}, \textcolor{orange}{girl}, \textcolor{orange}{aunt}, 
\textcolor{orange}{wife}, \textcolor{orange}{actress}, \textcolor{orange}{princess}, \textcolor{orange}{waitress}, \textcolor{orange}{sister}, \textcolor{orange}{queen}, \textcolor{orange}{pregnant}, \textcolor{orange}{daughter},  \textcolor{orange}{girlfriend}, \textcolor{orange}{chairwoman}, \textcolor{orange}{policewoman}, \textcolor{orange}{she}, \textcolor{orange}{her}, \textcolor{orange}{hers}, \textcolor{orange}{herself}, 
\textcolor{olive}{\textit{man}}, \textcolor{olive}{\textit{male}}, \textcolor{olive}{\textit{father}}, \textcolor{olive}{\textit{gentleman}}, \textcolor{olive}{\textit{boy}}, \textcolor{olive}{\textit{uncle}}, \textcolor{olive}{\textit{husband}}, \textcolor{olive}{\textit{actor}}, \textcolor{olive}{\textit{prince}}, \textcolor{olive}{\textit{waiter}}, \textcolor{olive}{\textit{son}}, \textcolor{olive}{\textit{brother}}, \textcolor{olive}{\textit{guy}}, \textcolor{olive}{\textit{emperor}}, \textcolor{olive}{\textit{dude}}, \textcolor{olive}{\textit{cowboy}}, \textcolor{olive}{\textit{boyfriend}}, \textcolor{olive}{\textit{chairman}}, \textcolor{olive}{\textit{policeman}}, \textcolor{olive}{\textit{he}}, \textcolor{olive}{\textit{his}}, \textcolor{olive}{\textit{him}}, \textcolor{olive}{\textit{himself}} and their plurals. 
\textcolor{orange}{Orange} denotes female words, whereas \textcolor{olive}{\textit{green}} denotes male words.
We select the gender words by investigating the datasets manually.

\paragraph{Answer distributions are skewed toward each gender.}

\begin{figure*}[h]
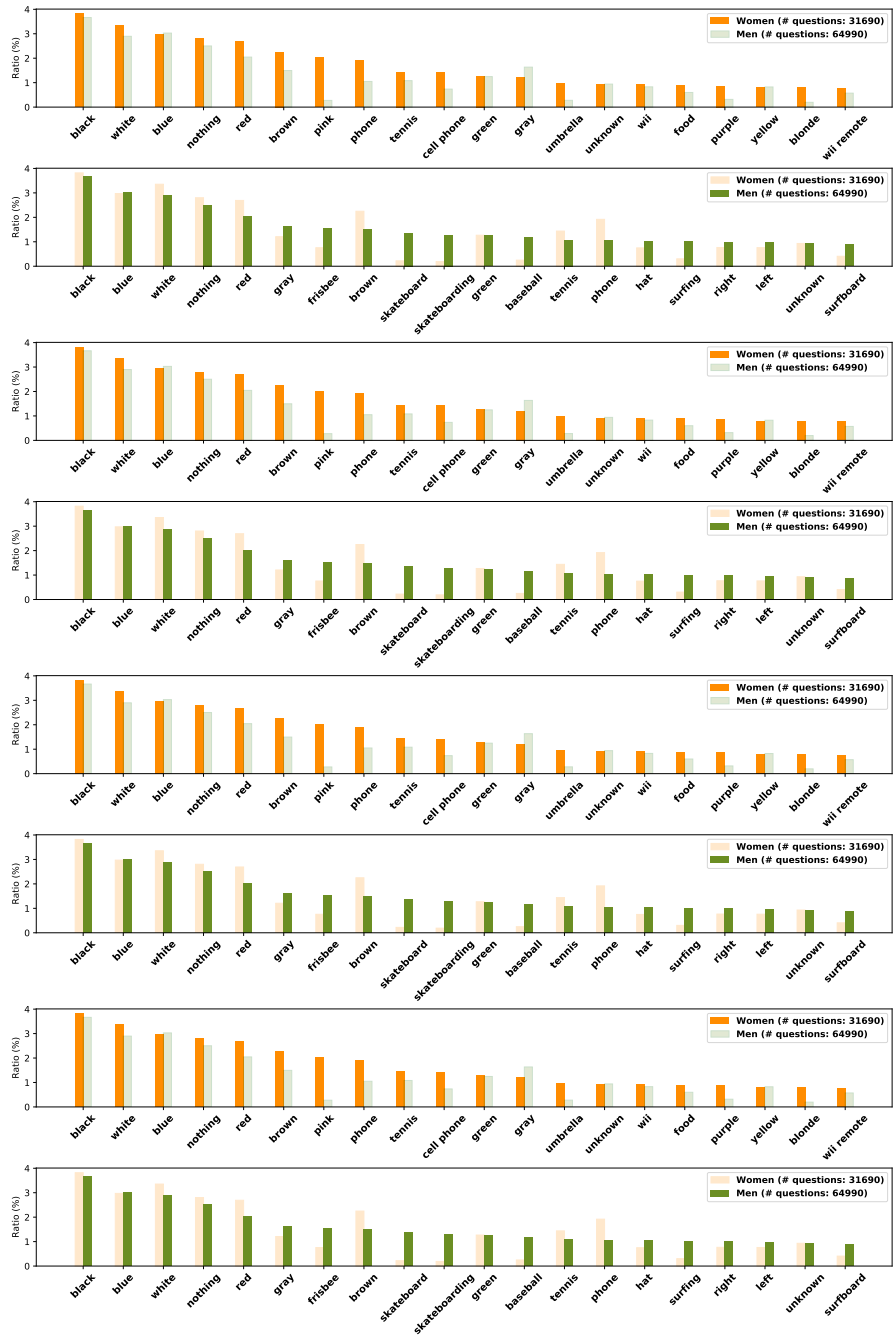

  \centering
  \includegraphics[scale=0.34]{follow_femtop.pdf}
  \includegraphics[scale=0.34]{follow_masctop.pdf}
  \includegraphics[scale=0.34]{follow_femtop.pdf}
  \includegraphics[scale=0.34]{follow_masctop.pdf}
  \includegraphics[scale=0.34]{follow_femtop.pdf}
  \includegraphics[scale=0.34]{follow_masctop.pdf}
  \includegraphics[scale=0.34]{follow_femtop.pdf}
  \includegraphics[scale=0.34]{follow_masctop.pdf}
  \caption{Top-20 frequent answers in GQA, Visual Genome, Visual7W, and OK-VQA:  Frequent answers for women questions (\textcolor{DarkOrange}{orange}). For the comparison, we also show the ratio of the answers over men questions (\textcolor{DarkGreen}{green}). \textbf{Below}: Frequent answers for men questions. As in above, we also show the ratio of answers for women questions. A large difference in ratio indicates that the answer is skewed toward certain gender.}
  \label{fig:sup-top20}
\end{figure*}

In the main paper, we show the top-$20$ answers of VQA 2.0 \cite{goyal2017making}. In the appendix, we also show the results of the other datasets (GQA \cite{hudson2019gqa}, Visual Genome \cite{krishna2016visual}, Visual7W \cite{zhu2016visual7w}, and OK-VQA \cite{marino2019ok}). As well as VQA 2.0, the distributions for women/men questions are skewed toward each gender in the other datasets.

\paragraph{Gender-answer correlations reflect gender stereotypes and discrimination.}
Here, we describe the detailed setting for BS (Section 4.3 in the main paper). We filter answers that do not appear more than $n$ times in women/men questions. For each dataset, we use: $n = 150$ (VQA 2.0), $n = 100$ (Visual Genome), $n = 100$ (GQA), $n = 10$ (Visual7W), $n = 5$ (OK-VQA).

\begin{figure*}[h]
  \centering
  \includegraphics[scale=0.34]{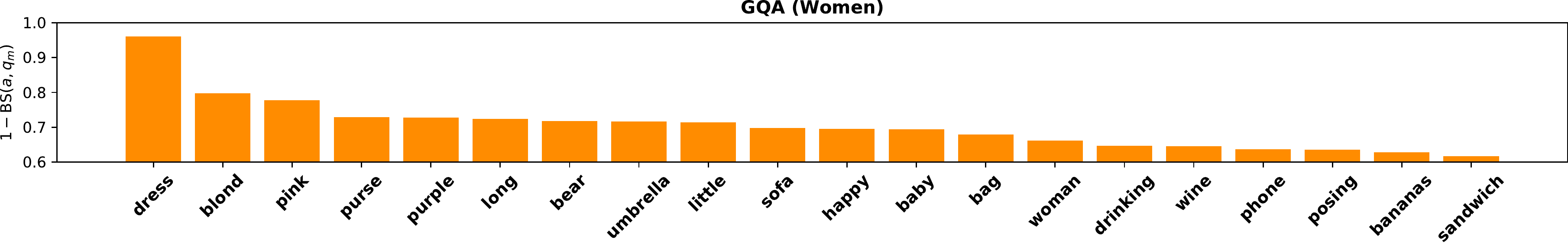}
  \includegraphics[scale=0.34]{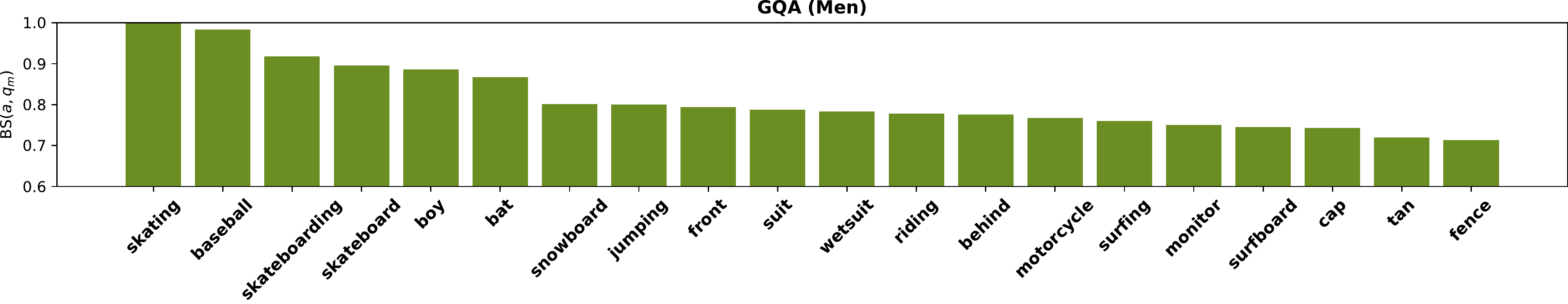}
  \includegraphics[scale=0.34]{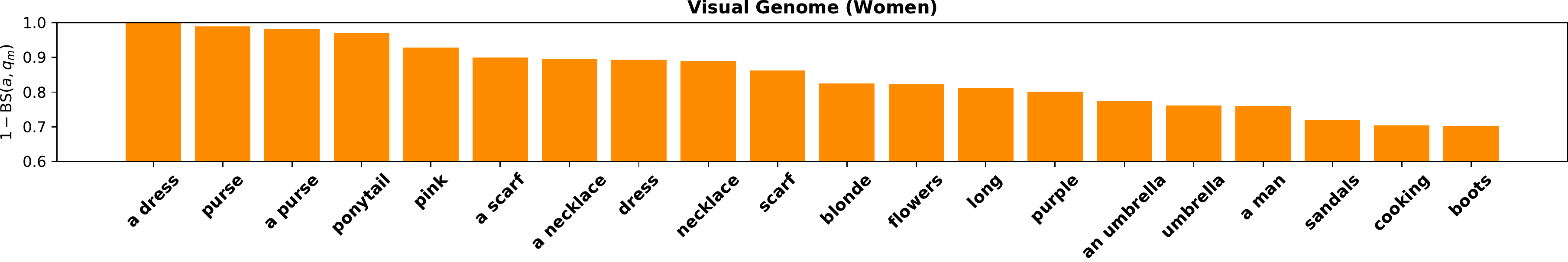}
  \includegraphics[scale=0.34]{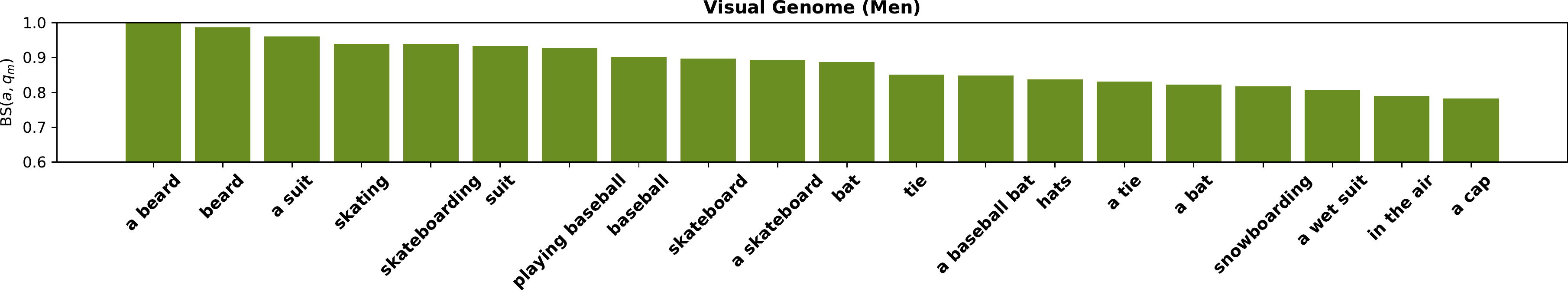}
  \includegraphics[scale=0.34]{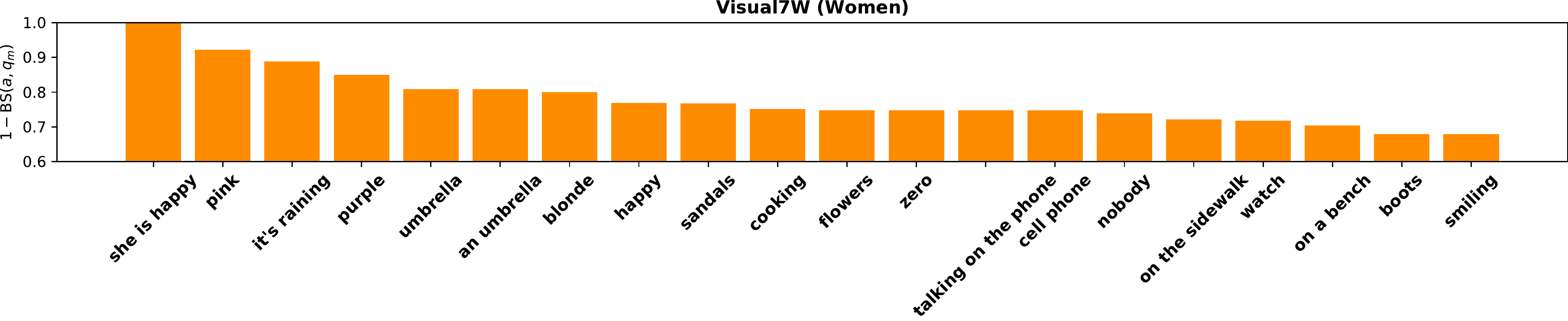}
  \includegraphics[scale=0.34]{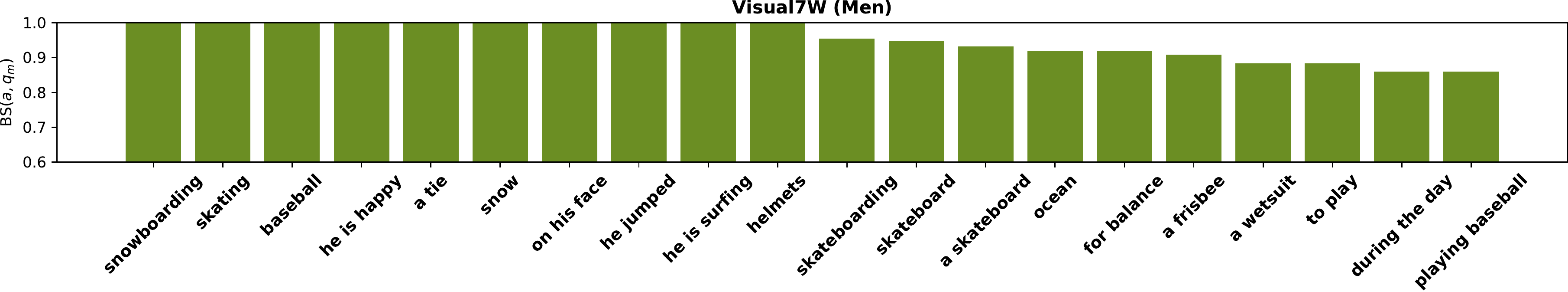}
  \includegraphics[scale=0.34]{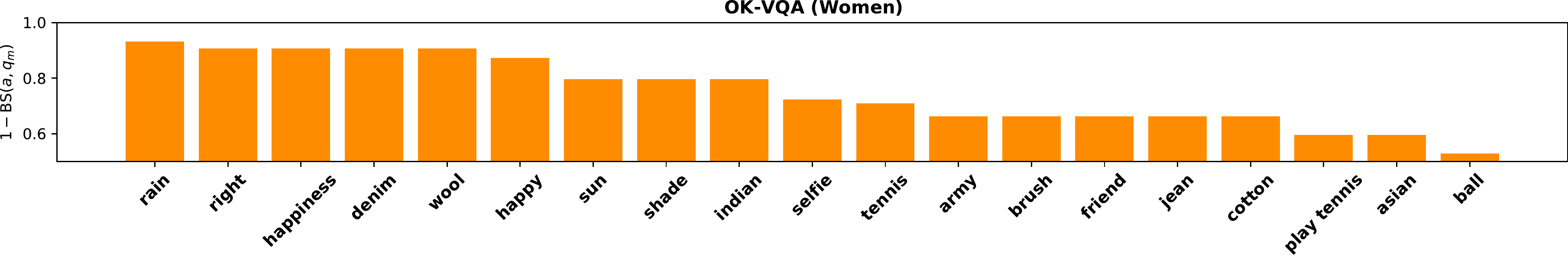}
  \includegraphics[scale=0.34]{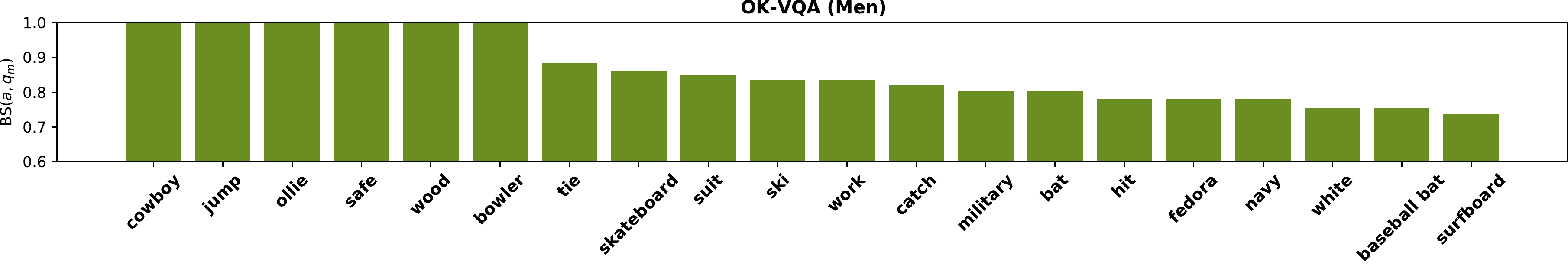}
  \caption{Top-20 answers that are co-related with women questions (\textbf{above}) and men questions (\textbf{below}) in GQA, Visual Genome, Visual7W, and OK-VQA.}
  \label{fig:sup-ans-gender}
\end{figure*}

We show the results of BS for GQA, Visual Genome, Visual7W, and OK-VQA in Figure~\ref{fig:sup-ans-gender}. As in the case of VQA 2.0 in the main paper, the distributions are highly skewed toward each gender.

\paragraph{Gender-stereotypical samples}

\begin{figure*}[h]
  \centering
  \includegraphics[width=\linewidth]{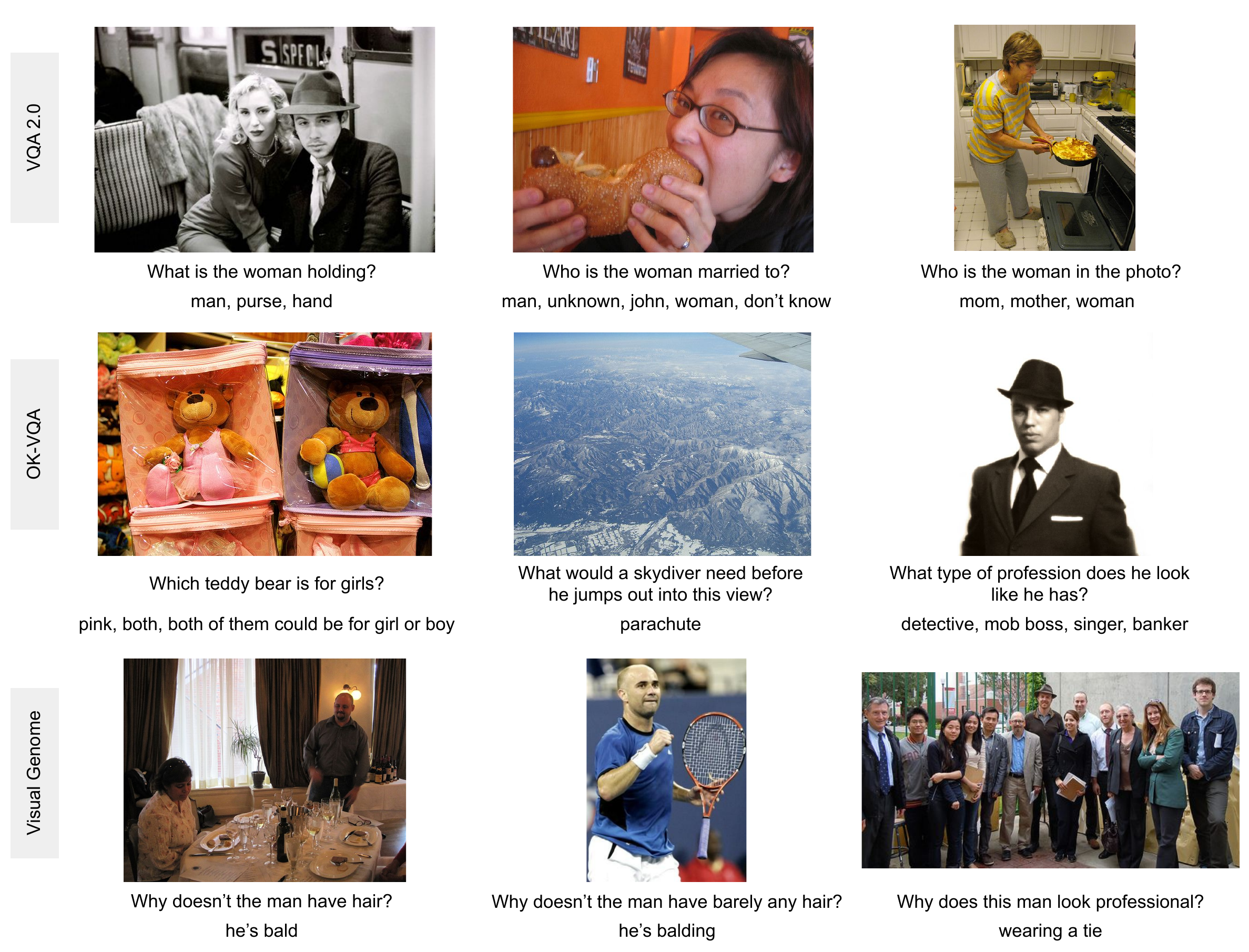}
  \caption{Examples of gender stereotypes in VQA 2.0 \cite{goyal2017making} (\textbf{above}), OK-VQA \cite{marino2019ok} (\textbf{middle}), and Visual Genome \cite{krishna2016visual} (\textbf{below})}
  \label{fig:sup-gender-sample}
\end{figure*}

In Figure~\ref{fig:sup-gender-sample}, we additionally show some samples that include harmful gender stereotypes in VQA 2.0 \cite{goyal2017making}, OK-VQA \cite{marino2019ok}, and Visual Genome \cite{krishna2016visual}.

\clearpage
\section{Racial Bias in VQA}

\paragraph{Racial-related words and nationality-related words}
We show the full list of racial-related words and nationality-related words used for the analysis in the paper. 

The list of racial-related words is: \textit{Black}, \textit{African}, \textit{Africa}, \textit{Latino}, \textit{Latina}, \textit{Latinx}, \textit{Hispanic}, \textit{White}, \textit{Caucasian}, \textit{Asian}, \textit{Oriental}, \textit{Asia}, \textit{Native}, \textit{Indigenous}, \textit{Arabic}. 

The list of nationality-related words is: \textit{American}, \textit{USA}, \textit{United States}, \textit{African American}, \textit{Chinese}, \textit{China}, \textit{Japanese}, \textit{Japan}, \textit{Indian}, \textit{India}, \textit{Mexican}, \textit{Mexico}, \textit{Italian}, \textit{Italy}, \textit{Spanish}, \textit{German}, \textit{French}, \textit{France}, \textit{English}, \textit{British}, \textit{England}, \textit{Russian}, \textit{Swiss},  \textit{Hawaiian}, \textit{Thai}, \textit{Brazil}.

\paragraph{Racial-stereotypical examples}

\begin{figure*}[h]
  \centering
  \includegraphics[width=\linewidth]{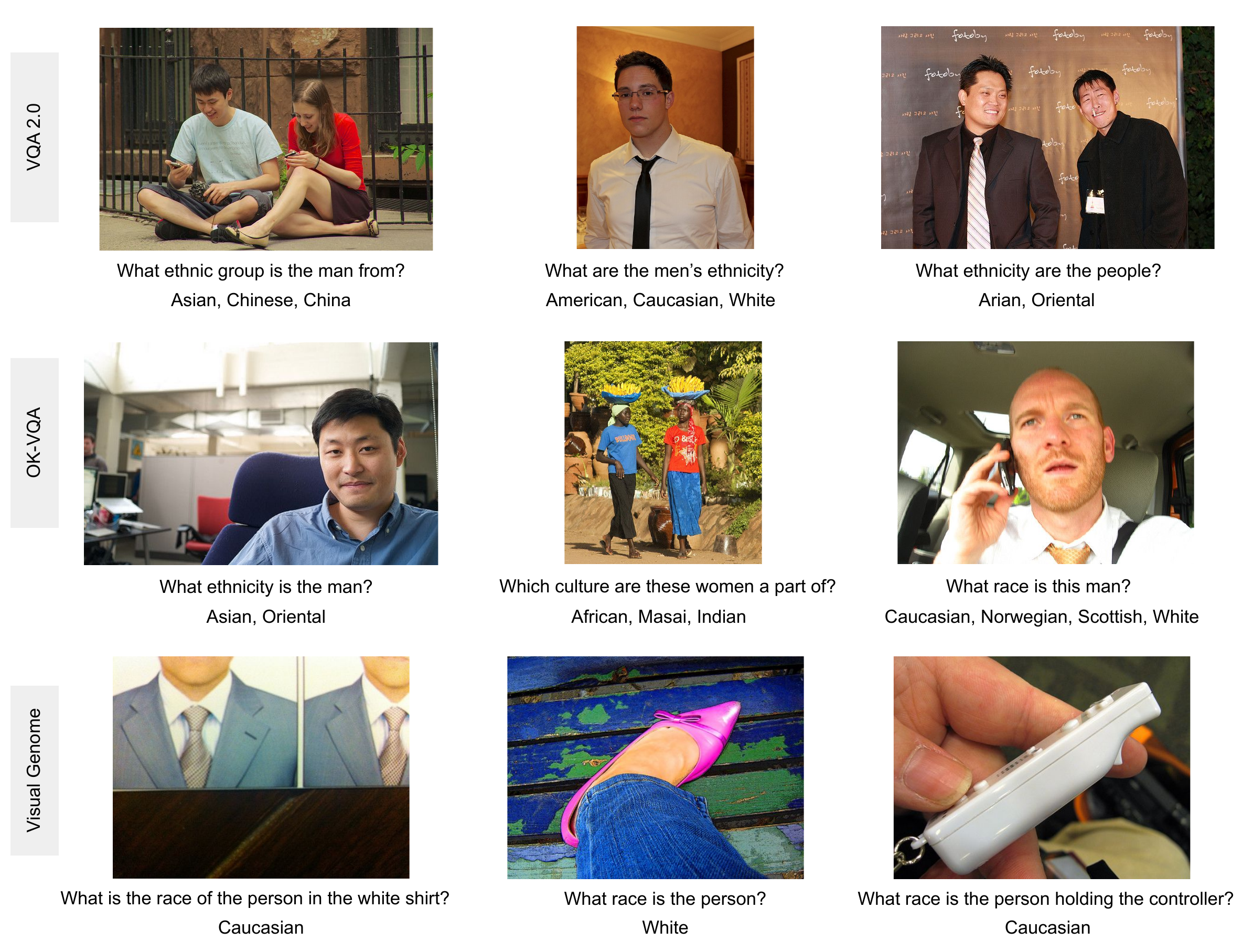}
  \caption{Examples of racial stereotypes in VQA 2.0 \cite{goyal2017making} (\textbf{above}), OK-VQA \cite{marino2019ok} (\textbf{middle}), and Visual Genome \cite{krishna2016visual} (\textbf{below})}
  \label{fig:sup-race-sample}
\end{figure*}

In Figure~\ref{fig:sup-race-sample}, we additionally show some samples that include harmful racial stereotypes in VQA 2.0 \cite{goyal2017making}, OK-VQA \cite{marino2019ok}, and Visual Genome \cite{krishna2016visual}.

\clearpage

%\bibliographystyle{ACM-Reference-Format}
%\bibliography{sample-base}

\end{document}